\documentclass[review]{elsarticle}

\usepackage{graphicx}
\usepackage{amsmath,amssymb} 
\usepackage{color}
\usepackage[titletoc,title]{appendix}

\usepackage{algorithm}
\usepackage{multirow}
\usepackage{algorithmic}
\usepackage{subfigure}
\usepackage{enumerate}

\newcommand\ie{\emph{i.e.}} 
 
\newcommand\etal{\emph{et al.}}

\begin{document}

\begin{frontmatter}

\title{Monocular Depth Estimation with Hierarchical Fusion of Dilated CNNs and Soft-Weighted-Sum Inference}

\author{Bo Li}
\author{Yuchao Dai}
\author{Mingyi He}
\address{Northwestern Polytechnical University, Xi'an, China}




\begin{abstract}
Monocular depth estimation is a challenging task in complex compositions depicting multiple objects of diverse scales. Albeit the recent great progress thanks to the deep convolutional neural networks (CNNs), the state-of-the-art monocular depth estimation methods still fall short to handle such real-world challenging scenarios. 

In this paper, we propose a deep end-to-end learning framework to tackle these challenges, which learns the direct mapping from a color image to the corresponding depth map. First, we represent monocular depth estimation as a multi-category dense labeling task by contrast to the regression based formulation. In this way, we could build upon the recent progress in dense labeling such as semantic segmentation. Second, we fuse different side-outputs from our front-end dilated convolutional neural network in a hierarchical way to exploit the multi-scale depth cues for depth estimation, which is critical to achieve scale-aware depth estimation. Third, we propose to utilize soft-weighted-sum inference instead of the hard-max inference, transforming the discretized depth score to continuous depth value. Thus, we reduce the influence of quantization error and improve the robustness of our method. Extensive experiments on the NYU Depth V2 and KITTI datasets show the superiority of our method compared with current state-of-the-art methods. Furthermore, experiments on the NYU V2 dataset reveal that our model is able to learn the probability distribution of depth. 

\end{abstract}


\end{frontmatter}


\section{Introduction}
Depth estimation aims at predicting pixel-wise depth for a single or multiple images, which is an essential intermediate component toward 3D scene understanding. It has been shown that depth information can benefit tasks such as recognition~\cite{ren2012rgb,shotton2013real}, human computer interaction~\cite{fanello2014learning}, and 3D model reconstruction~\cite{Silberman2012Indoor}. Traditional techniques have predominantly worked with multiple images to make the problem of depth prediction well-posed, which include $N$-view reconstruction, structure from motion (SfM) and simultaneous localization and mapping (SLAM) \cite{}.

However depth estimation from a monocular single viewpoint lags far behind its multi-view counterpart. This is mainly due to the fact that the problem is ill-posed and inherently ambiguous: a single image on its own does not provide any depth cue explicitly (\ie, given a color image of a scene, there are infinite number of 3D scene structures explaining the 2D measurements exactly). When specific scene dependent knowledge is available, depth estimation or 3D reconstruction from single images can be achieved by utilizing geometric assumptions such as the ``Blocks World'' model \cite{Block-World:ECCV-2010}, the ``Origami World'' model \cite{Origami-World:ECCV-2014}, shape from shading [7] and repetition of structures [8]. However, these cues typically work for images with specific structures and may not be applied to general scenarios.

Recently, learning based monocular depth estimation methods that predicting scene geometry directly by learning from data, have gained popularity. Typically, such approaches recast the underlying depth estimation problem in a pixel-level scene labeling pipeline by exploiting relationship between monocular image and depth.  Fully-convolutional neural network has been proved to be an effective method to solve these kinds of problems. There have been considerable progress in applying deep convolutional neural network (CNN) to this problem and excellent performances have been achieved \cite{li2015depth,liu2015deep, wang2015towards,eigen2015predicting,cao2016estimating,laina2016deeper,Xu2017Multi,Godard2017Unsupervised}.

Albeit the above success, state-of-the-art monocular depth estimation methods still fall short to handle real world challenging complex decompositions depicting multiple objects of diverse scales due to the following difficulties: 1) the serious data imbalance problem due to the perspective effect,  where samples with small depths are much more than samples with large depths; 2) there are more rapid changes in depth value compared with other dense predictions tasks such as semantic labeling and 3) long range context information is needed handle the scale ambiguity in depth estimation. Even though there have been various post-processing methods to refine the estimated depth from the deep network map~\cite{li2015depth,liu2015deep, wang2015towards,eigen2015predicting,cao2016estimating, laina2016deeper,Xu2017Multi,Godard2017Unsupervised}, the bottleneck in improving monocular depth estimation is still the specially designed CNN architecture, which is highly desired. 



\begin{figure*}[ht]
\centering
\includegraphics[width=0.9\linewidth]{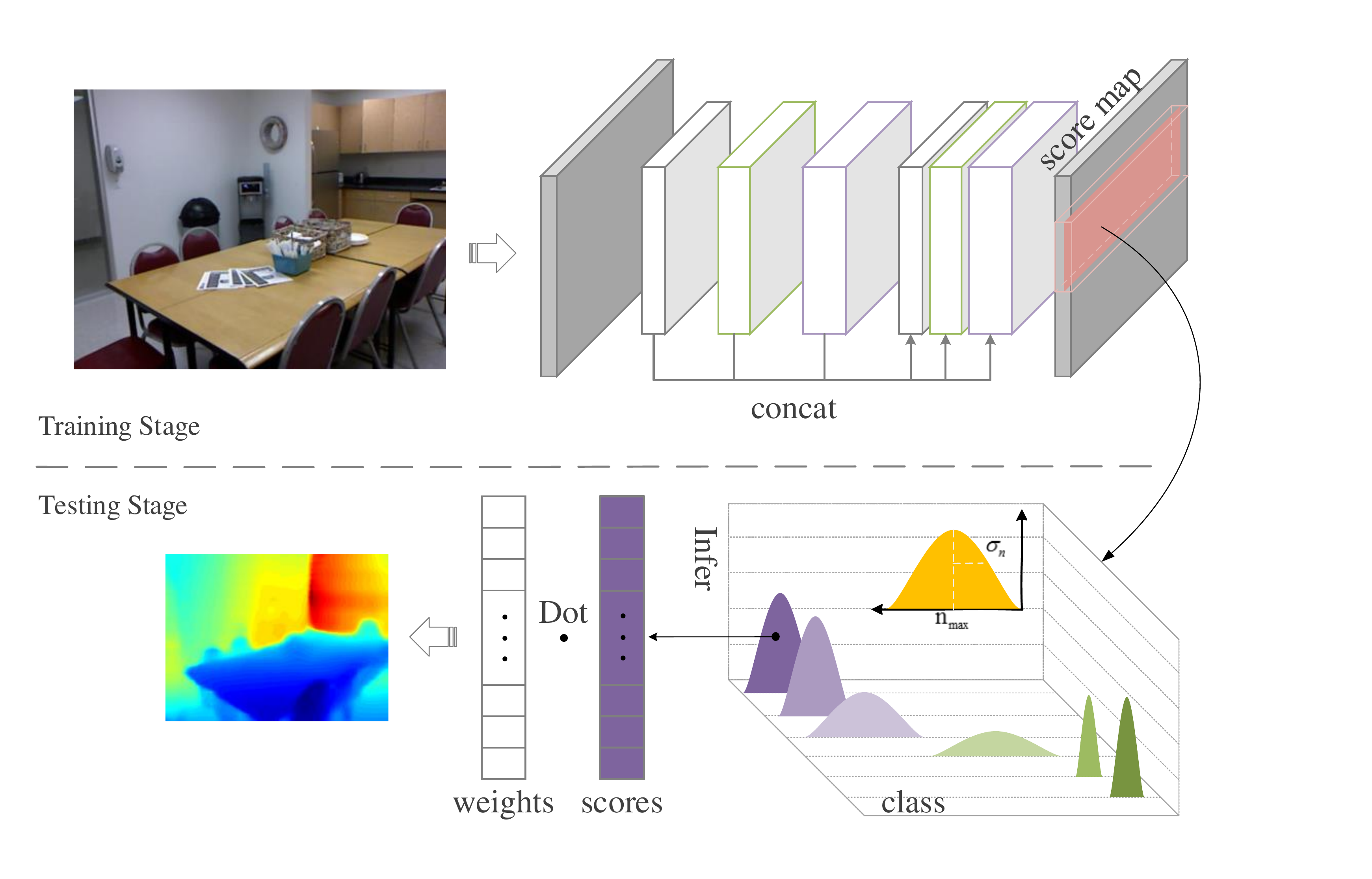}
\caption{Flowchart of our monocular depth estimation framework, which is built upon deep Residual network \cite{ResNet} and consists of dilated convolution and hierarchical feature fusion. Soft-weighted-sum inference is used to predict continuous depth values from the discrete depth labels. We also illustrate typical probability distribution of labels from the network, which shows that our classification based framework is able to learn the similarity between labels.}
\label{fig:flowchart}
\end{figure*}

In this paper, we present a deep CNN based framework to tackle the above challenges, which learns the direct mapping from the color image to the corresponding depth map in an end-to-end manner. We recast monocular depth estimation as a multi-category dense labeling as contrast to the widely used regression formulation. Our network is based on the deep residual network~\cite{ResNet}, where dilated convolution and hierarchical fusion layers are designed to expand the receptive field and to fuse multi-scale depth cues. In order to reduce the influence of quantization error and improve the robustness of our method, we propose to use a soft-weighted-sum inference. Extensive experimental results show that even though we train our network as a standard classification task with the multinomial logistic loss, our network is able to learn the the probability distribution among different categories. The overall flowchart of our framework is illustrated in Fig.~\ref{fig:flowchart}.

Our main contributions can be summarized as:
\begin{itemize}
\item We propose a deep end-to-end learning framework to monocular depth estimation by recasting it as a classification task, where both dilated convolution and hierarchical feature fusion are used to learn the scale-aware depth cues. 
\item Our network is able to output the probability distribution among different depth labels. We propose a soft-weighted-sum inference, which could reduce the influence of quantization error and improve the robustness.
\item Our method achieves the state-of-the-art performance on both indoor and outdoor benchmarking datasets, NYU V2 and KITTI dataset.
\end{itemize}

\section{Related work}

In this section, we briefly review related works for monocular depth estimation, which can be roughly categories as conventional MRF/CRF based methods and deep learning based methods. 

\textbf{MRF/CRF Based Methods:} Seminal work by Saxena \etal~\cite{saxena2009make3d,saxena2007depth} tackles the problem with a multi-scale Markov Random Field (MRF) model, with the parameters of the model learned through supervised learning. Liu \etal~\cite{liu2010single} estimated the depth map from predicted semantic labels, achieving improved performance with a simpler MRF model. Ladicky \etal~\cite{ladicky2014pulling} showed that perspective geometry can be used to improve results and demonstrated how scene labeling and depth estimation can benefit each other under a unified framework, where a pixel-wise classifier was proposed to jointly predict a semantic class and a depth label from a single image. Besides these parametric methods, other works such as~\cite{karsch2012depth,liu2014discrete,konrad20122d} recast monocular depth estimation in a non-parametric fashion, where the whole depth map is inferred from candidate depth maps. Liu \etal~\cite{liu2014discrete} proposed a discrete-continue CRFs, which aims to avoid the over-smoothing and maintain occlusion boundaries. Anirban \etal~\cite{} proposed a Neural Regression Forest model for this problem. These works provide important insights and cues for single image depth estimation problem, while most of them utilized the hand-crafted features thus limited their performance especially for complex scenarios.


\textbf{Deep Learning Based Methods:} Recently, monocular depth estimation has been greatly advanced thanks to deep convolutional neural network (CNN). Eigen \etal~\cite{eigen2014depth} presented a framework by training a large hierarchical deep CNN. However, partly due to the fully connect layers used in the network model, their network have to be trained with very large scale data. By contrast, Li \etal~\cite{li2015depth} proposed a patch-based CNN framework and a hierarchical-CRF model to post-process the raw estimated depth map, which significantly reduces the number of training image needed. Liu \etal~\cite{liu2015deep} proposed a CRF-CNN joint training architecture, which could learn the parameters of the CRF and CNN jointly. Wang \etal~\cite{wang2015towards} proposed a CNN architecture for joint semantic labeling and monocular depth prediction. Chen \etal~\cite{Chen2017Single} proposed an algorithm to estimate metric depth using annotations of relative depth.

Very recently, Laina \etal~\cite{laina2016deeper} proposed using the Huber loss instead of the $L_2$ loss to deal with the long tail effect of the depth distribution. Cao \etal~\cite{cao2016estimating} demonstrated that formulating depth estimation as a classification task could achieve better results than regression with $L_2$ loss, while insufficient analysis is given for the success. In addition, different with our method, they used hard-max inference in the testing phase. Xu \etal~\cite{Xu2017Multi} proposed a Multi-Scale Continuous CRFs to better extract the hierarchical information and improve the smoothness of the final results. Our hierarchical information fusion strategy is much simpler than~\cite{Xu2017Multi}, while we also achieve comparable results. 


\textbf{Unsupervised monocular depth learning} Besides the above methods using ground truth depth maps to supervise the network learning, there is another group of methods that using novel view synthesis to supervised the network learning by exploiting the availability of stereo images and image sequences \cite{Garg2016Unsupervised} \cite{Godard2017Unsupervised} \cite{Unsupervised-Depth-Motion} \cite{Semi-supervised-Depth-CVPR-17} cite{Unsupervised-Depth-Motion}. Garg \etal \cite{Garg2016Unsupervised} proposed to train a network for monocular depth estimation using an image reconstruction loss, where a Taylor approximation is performed to linearize the loss. Godard \etal \cite{Godard2017Unsupervised} replaced the use of explicit depth data during training with easier-to-obtain binocular stereo footage, which enforces consistency between the disparities produced relative to both the left and right images, leading to improved performance and robustness compared to existing approaches. Along this pipeline, Zhou \etal \cite{Unsupervised-Depth-Motion} presented an unsupervised learning framework for the task of monocular depth and camera motion estimation from unstructured video sequences based on image warping to evaluate the image error. Kuznietsov \etal \cite{Semi-supervised-Depth-CVPR-17} learned depth in a semi-supervised way, where sparse ground-truth depth and photoconsistency are jointly used. Ummenhofer \etal \cite{DeMon-CVPR-2017} trained a convolutional network end-to-end to compute depth and camera motion from successive, unconstrained image pairs, where the architecture is composed of multiple stacked encoder-decoder networks. 

The key supervision signal for these ``unsupervised'' methods comes from the task of novel view synthesis: given one input view of a scene, synthesize a new image of the scene seen from a different camera pose. Essentially, pairs of rectified stereo images or consecutive image frames have already encode the depth information implicitly.

Our work is also related to the works on FCN (fully convolutional network) based dense labeling. Long \etal~\cite{Long2014Fully} proposed the fully convolution neural network for semantic segmentation, which is widely used in other dense labeling problems. Hariharan \etal~\cite{Hariharan2015Hypercolumns} presented that low-level CNN feature is better to the boundary preserving and object location. Recently, Yu \etal~\cite{yu2015multi} demonstrated that dilated convolution could expand the receptive field of the corresponding neuron while keeping the resolution of the feature map. Chen~\cite{Chen2014Semantic} successfully utilized the dilated convolution on the semantic problem and show how to build them on the pre-trained CNN.

\section{Our Framework}
Targeting at handling the real world challenges with the current state-of-the-art methods, we propose a deep end-to-end learning framework to monocular depth estimation, which learns the direct mapping from a color image to the corresponding depth map. Our framework to monocular depth estimation consists of two stages: model training with classification loss and inference with soft-weighted sum. First, by recasting monocular depth estimation as multi-class labeling, we design an hierarchical fusion dilated CNN to learn the mapping from an RGB image to the corresponding depth score map directly. Our network architecture hierarchically fuses multi-scale depth features, which is important to achieve scale-aware monocular depth estimation. Second, we propose a soft-weighted-sum inference as contrast to the hard-max inference, which transfers the discretized depth scores to continuous depth values. In this way, we could reduce the influence of quantization error and improve the robustness.

\subsection{Network Architecture}
\begin{figure*}[ht]
\centering{\includegraphics[width=0.9\linewidth]{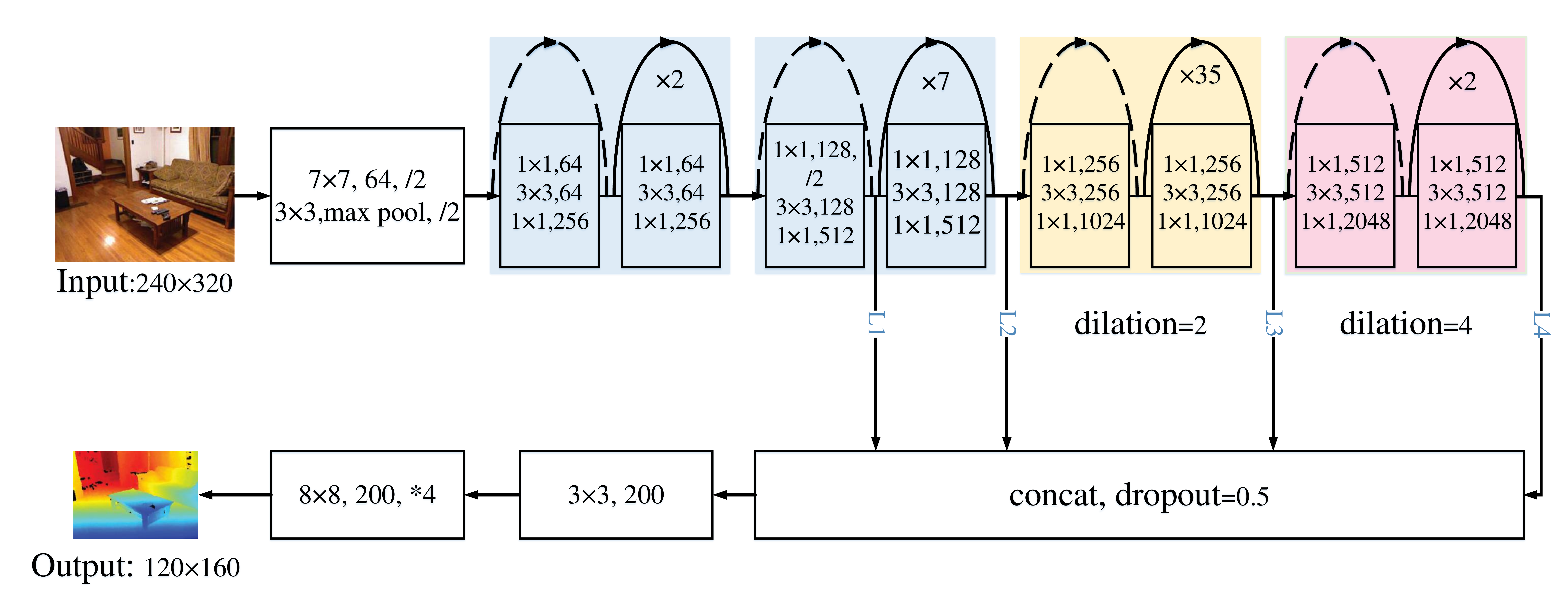}}
\caption{Illustration of our network architecture. The detail of the basic residual block could be referred to~\cite{ResNet}. $\times n$ means the block repeats $n$ times. We present all the hyper-parameters of convolution and pooling layers. All the convolution layers are followed by batch normalization layer except for the last one. $/2$ means the layer's stride is 2. $*4$ means the deconv layer's stride is 4. Dilation shows the dilated ratio of the corresponding parts. $L1, \cdots L6$ are our skip connection layers.}
\label{network}
\end{figure*}

Our CNN architecture is illustrated in Fig.~\ref{network}, in which the weights are initialized from a pre-trained 152 layers deep residual CNN (ResNet) \cite{ResNet}. Different from existing deep network \cite{VGGNet}, ResNet \cite{ResNet} explicitly learns residual functions with reference to the layer inputs, which makes it easier to optimize with higher accuracy from considerably increased network depth. ResNet \cite{ResNet} was originally designed for image classification. In this work, we re-purpose it to make it suitable to our depth estimation task by 
\begin{itemize}
\item Removing all the fully connect layers. In this way, we greatly reduce the number of model parameters as most of the parameters are in the fully connect layers \cite{eigen2015predicting}. Although preserving the fully connect layers is beneficial to extract long range context information, our experiments show that it is unnecessary in our network thanks to dilated convolution.

\item Taking advantage of the dilated convolution \cite{yu2015multi}. Dilated convolution could expand the receptive field of the neuron without increasing the number of model parameters. Furthermore, with the dilated convolution, we could remove some pooling layers without decreasing the size of receptive field of correspondent neurons. In addition, we could keep the resolution of the feature map and final results, \ie, the output resolution has been increased.

\item Hierarchal fusion. We concatenate intermediate feature maps with the final feature map directly. This skip connection design could benefit the multi-scale feature fusion and boundary preserving. 
\end{itemize}

\begin{figure}[htb]
\centering
\scalebox{0.9}{
\begin{tabular}{@{}c@{}c@{}c}
\includegraphics[width=0.3\linewidth]{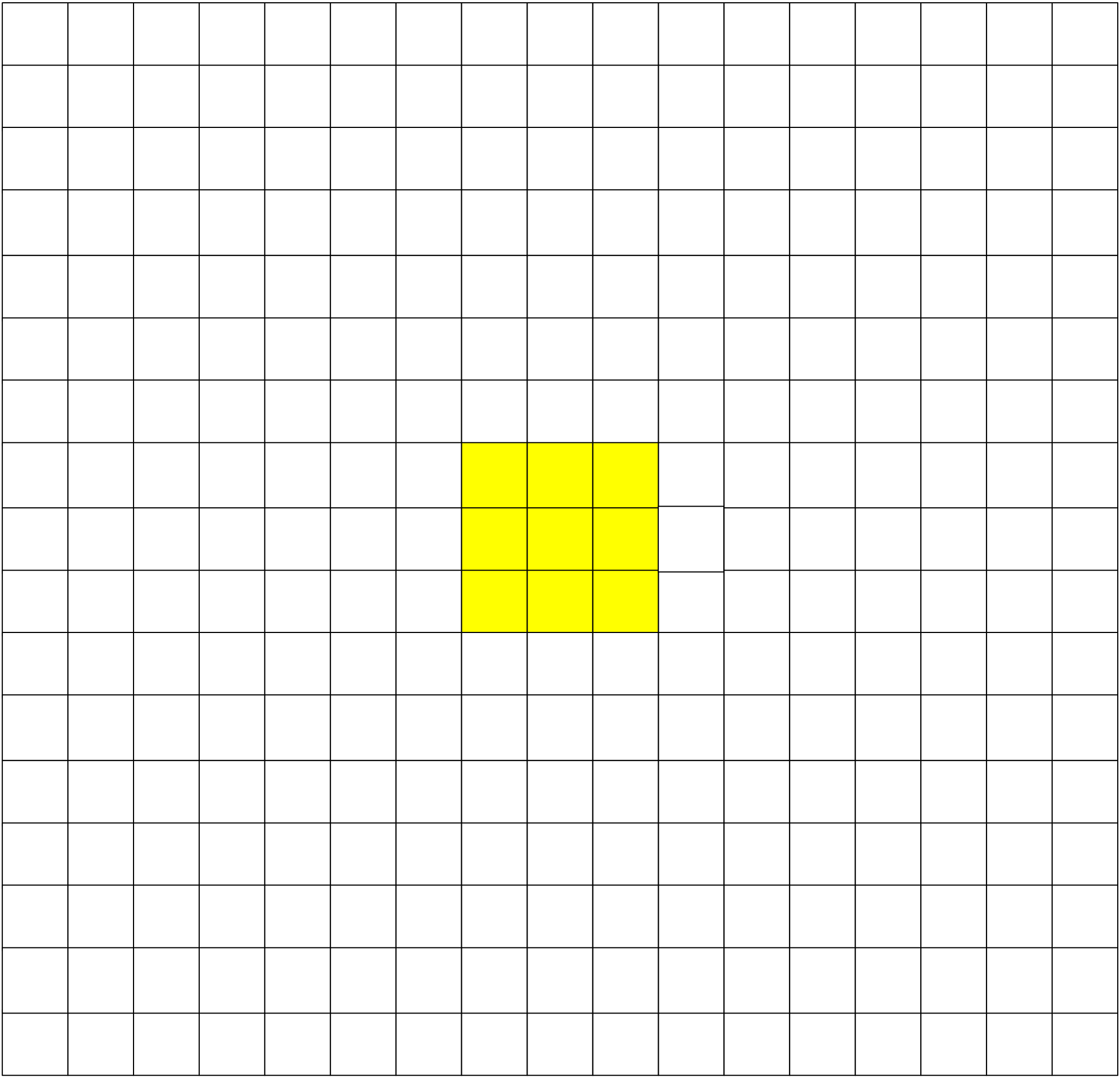}\ \ \ \ \ &\includegraphics[width=0.3\linewidth]{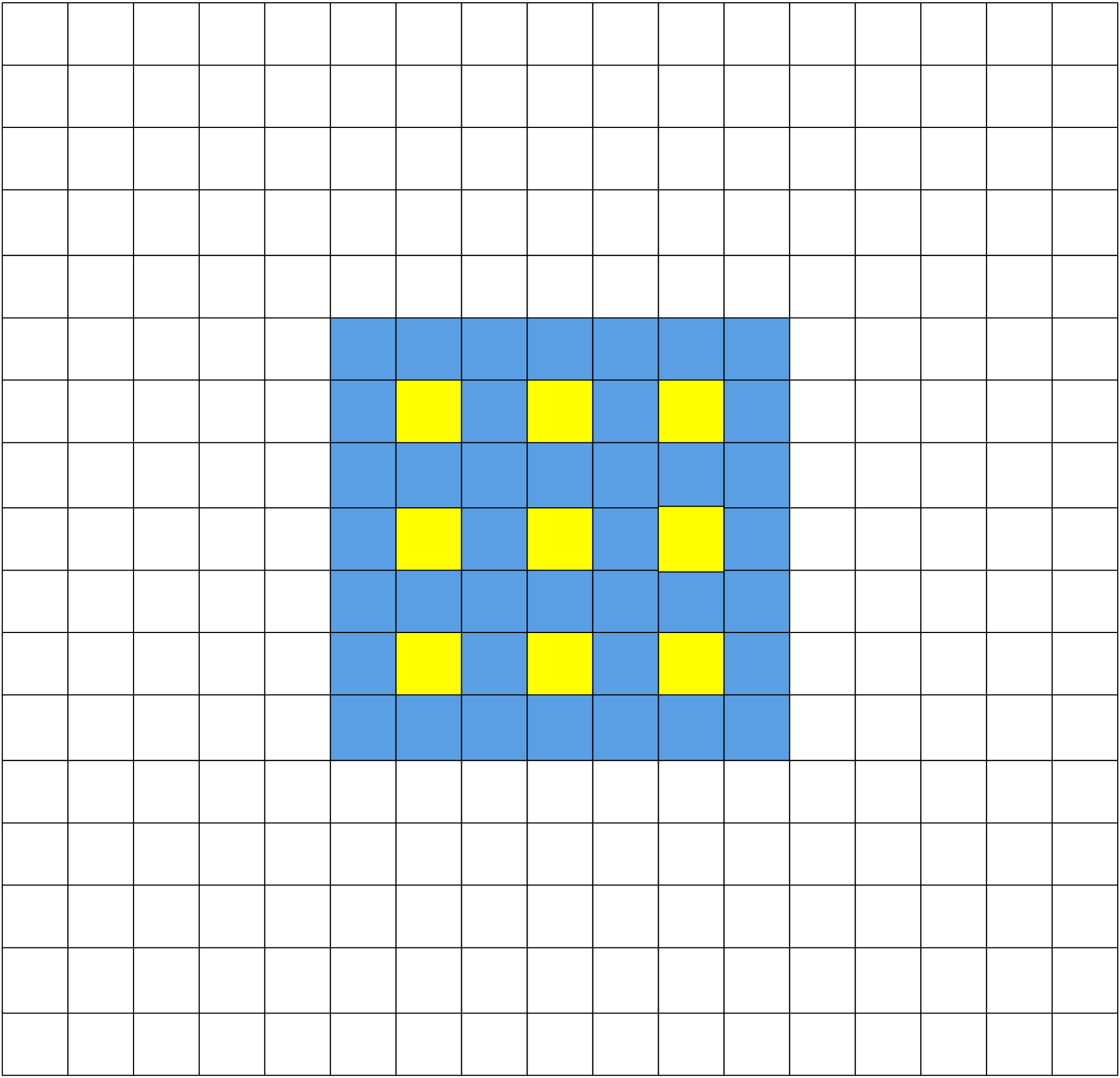}\ \ \ \ \
&\includegraphics[width=0.3\linewidth]{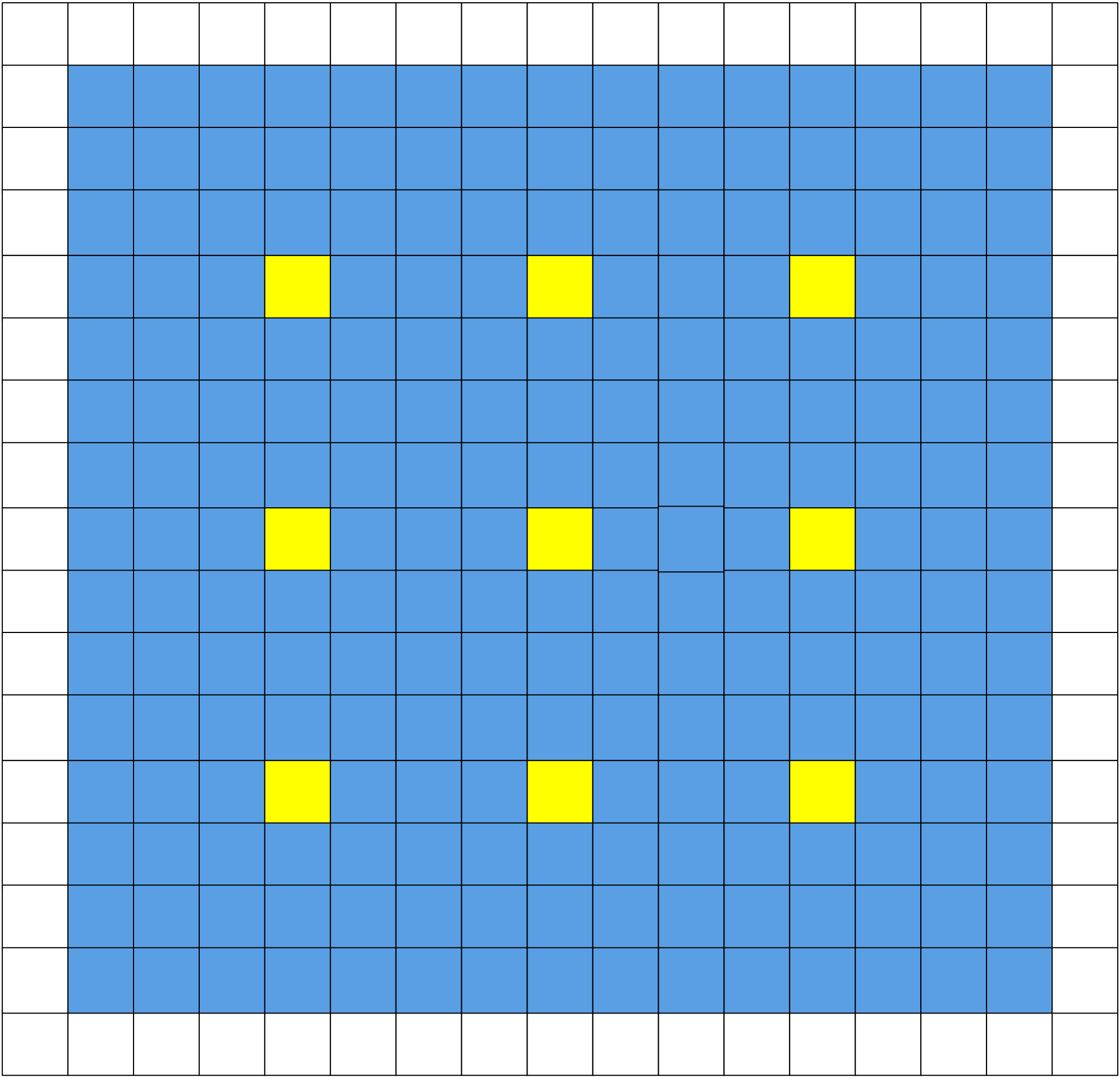}\\
(a) & (b) & (c)\\
\end{tabular}
}
\caption{Systematic dilation supports exponential expansion of the receptive field without loss of resolution or coverage. (a), (b), (c) are 1-dilated, 2-dilated, 4-dilated convolution respectively. And the corresponding receptive fields are $3\times3$, $7\times7$, and $15\times15$. The receptive field grows exponentially while the number of parameters is fixed.}
\label{fig:dilation}
\end{figure}

\textbf{Dilated Convolution}: Recently, dilated convolution \cite{yu2015multi} has been successfully utilized in deep convolutional neural network, which could expands the field of perception without increasing the number of model parameters. Specially, let $F : \mathcal{Z}^2 \rightarrow \mathcal{R}$ be a discrete function. Let $\Omega_r = \left[ r,r\right] ^2 \cap \mathcal{Z} ^2 $ and let $ k: \Omega_r \rightarrow \mathcal{R}$ be a discrete filter of size $ \left( 2r+1 \right)^2  $. The discrete convolution filter $*$ can be expressed as
\begin{equation}
\label{convolution}
( F * k ) (\mathbf{p}) = \sum\limits_{\mathbf{s} + \mathbf{t} = \mathbf{p}}F(\mathbf{s})k(\mathbf{t}).
\end{equation}
We now generalize this operator. Let $l$ be a dilation factor and let $\mathbf{*}_l$ be defined as
\begin{equation}
\label{dilated convolution}
( F *_l k ) (\mathbf{p}) = \sum\limits_{\mathbf{s} + l\mathbf{t} = \mathbf{p}}F(\mathbf{s})k(\mathbf{t}).
\end{equation}
We refer to $ \mathbf{*}_l $ as a dilated convolution or an $l$-dilated convolution. The conventional discrete convolution $*$ is simply the 1-dilated convolution. An illustration of dilated convolution could be found in Fig.~\ref{fig:dilation}.

\textbf{Hierarchical Fusion:} As the CNN is of hierarchical structure, which means high-level neurons have larger receptive field and more abstract features, while the low-level neurons have smaller receptive field and more detail information.  Thus, combining multi-scale informations for pixel-level prediction tasks have received considerable interests. 


We propose to concatenate the high-level feature map and the intermediate feature map. The skip connection structure benefits both the multi-scale fusion and boundary preserving. In our network, the $L1, L2, L3, L4$ layers are of the same size, we concatenate them directly.

%

In conclusion, we briefly summarize our final network design. Typically, the pre-trained residual network is consisted of 4 parts. We remove the max-pooling layer in the last 2 parts and expand the corresponding convolution kernel with dilation 2 and 4 respectively. Then, a concatenation layer is added to fuse the hierarchical multi-scale informations from layers $L1-L4$. The last two layers of our network are convolution layer and deconvolution layer. The parameters setting is presented in Fig.~\ref{network}.

\subsection{Soft-Weighted-Sum Inference}

We reformulate depth estimation as classification task by equally discretizing the depth value in log space. Specifically,
\begin{equation}
\label{discrete}
l = round((\log(d) - \log(d_{min})) / q),
\end{equation}
where $l$ is the quantized label, $d$ is the continuous depth value, $d_{min}$ is the minimum depth value in the dataset or set to be a small value like $0.1$. $q$ is the width of the quantization bin.

With the quantization label, we train our network with the multinomial logistic loss.
\begin{equation}
\label{logistic loss}
L(\theta) = -\left[\sum_{i=1}^N \sum_{k=1}^K 1\{ y^{(i)}=k\} \log \frac{\exp(\theta^{(k)T}x^{(i)})}{\sum_{i=1}^K \exp( \theta^{(j)T} x^{(i)})}\right],
\end{equation}
where $N$ is the number of training samples, $\exp(\theta^{(k)T}x^{(i)})$ is the probability of label $k$ of sample $i$, and $k$ is the ground truth label.


In the testing stage, we propose to use the soft-weighted-sum inference. It is worth noting that, this method transforms the predicted score to the continuous depth value in a natural way. Specifically:
\begin{equation}
\label{soft-weight-sum}
\hat{d} = \exp\{ \mathbf{w}^{T} \mathbf{p} \}, w_i = \log(d_{min}) + q * i,
\end{equation}
where $ \mathbf{w} $ is the weight vector of depth bins. $\mathbf{p} $ is the output score. In our experiments, we set the number of bins to 200.


%
%


\subsection{Data Augmentation}
Although the training dataset is of ten thousands images, we still find the data augmentation is important to improve the final performance. In this work, we augment our dataset by 4 times for both the NYU v2 and the KITTI dataset. The augmentation methods we utilized include:
\begin{itemize}
  \item Color: Color channels are multiplied by a factor $c\in[0.9,1.1]$ randomly.
  \item Scale: We scale the input image by a factor of $s\in[1.3,1.5]$ randomly and crop the center patch of images to match the network input size.
  \item Left-Right flips: We flip left and right images horizontally.
  \item Rotation: We rotate the input image randomly by a factor of $r\in[-5,5]$.
\end{itemize}

\subsection{Implementation details}
Before proceeding to the experimental results, we give implementation details of our method. Our implementation is based on the efficient CNN toolbox: caffe~\cite{jia2014caffe} with an NVIDIA Tesla Titian X GPU.

The proposed network is trained by using stochastic gradient decent with batch size of 1 (This size is too small, thus we average the gradient of 8 iterations for one back-propagation), momentum of 0.9, and weight decay of 0.0004. Weights are initialized by the pre-trained model from ResNet \cite{ResNet}. The network is trained with iterations of 50k by a fixed learning rate 0.001 in the first 30k iterations, then divided by 10 every 10k iterations.
%
%
\section{Experimental Results}
In this section, we report our experimental results on monocular depth estimation for both outdoor and indoor scenes. We used the NYU V2 dataset, and the KITTI dataset, as they are the the largest open dataset we can access at present. We compared our method with the state-of-the-art methods published recently.

For quantitative evaluation, we report errors obtained with the following metrics, which have been extensively used in~\cite{saxena2009make3d, liu2010single, eigen2014depth, ladicky2014pulling,liu2014discrete}.
\begin{itemize}
\item Threshold: $\%$ of $d_i$  s.t. $ \max\left(\frac{\hat{d}_i}{d_i},\frac{d_i}{\hat{d}_i}\right)=\delta<thr$
\item Mean relative error (Rel): $\frac{1}{\vert T \vert}\sum_{d \in T}\vert{\hat{d}-d}\vert/d$
\item Mean $\mbox{log}_{10}$ error ($\mbox{log}_{10}$): $\frac{1}{\vert T \vert}\sum_{d \in T}\vert{\log_{10}\hat{d}-\log_{10}d}\vert$
\item Root mean squared error (Rms): $\sqrt{\frac{1}{\vert T \vert}\sum_{d \in T}{\Vert{\hat{d}-d}\Vert}^2}$
\end{itemize}
where $d$ is the ground truth depth, $\hat{d}$ is the estimated depth, and $T$ denotes the set of all points in the images.

\subsection{NYU V2 dataset}
The NYU V2 dataset~\cite{Silberman2012Indoor} contains around 240k RGB-depth image pairs, of which comes from 464 scenes, captured with a Microsoft Kinect. The official split consists of 249 training and 215 testing scenes. We equally sampled frames out of each training sequence, resulting in approximately 24k unique images. After off-line augmentations, our dataset comprises of approximately 96k RGB-D image pairs. We fill in the invalid pixels of the raw depth map with the ``colorization'' method, which is provided in the toolbox of NYU V2 dataset~\cite{Silberman2012Indoor}.

The original image resolution is $480 \times 640$. We downsampled the images to $240 \times 320$ as our network input. The resolution of the our network output is $120 \times 160$, which is half of the input size. In this dataset, we quantize the depth value into 200 bins. 

In Table~\ref{tab:nyu2}, we compared our method with the recent published state-of-the-art methods~\cite{eigen2015predicting,cao2016estimating,laina2016deeper,Xu2017Multi}.

\begin{figure*}[htb]
\centering
\scalebox{0.9}{
\begin{tabular}{@{}c@{}c@{}c@{}c@{}c@{}c@{}c}
(a) \ \
& \includegraphics[width=0.15\linewidth]{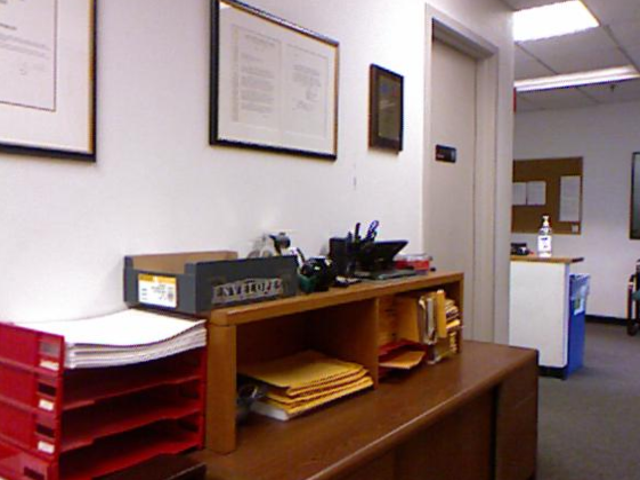} \ \
& \includegraphics[width=0.15\linewidth]{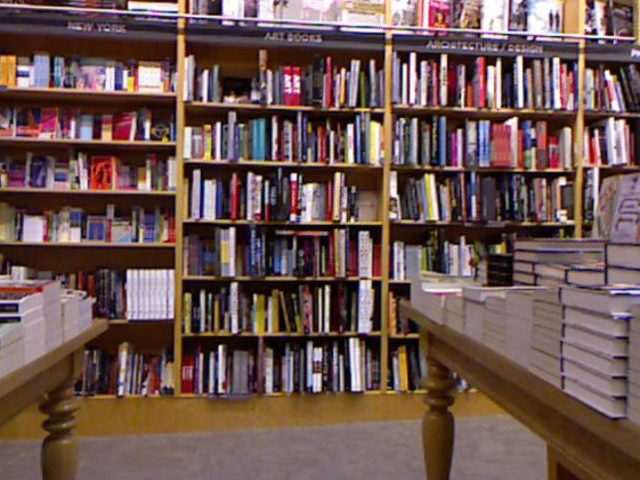} \ \
& \includegraphics[width=0.15\linewidth]{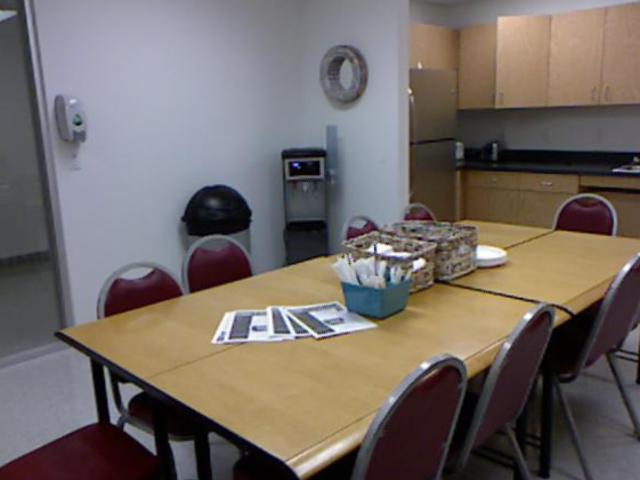} \ \
& \includegraphics[width=0.15\linewidth]{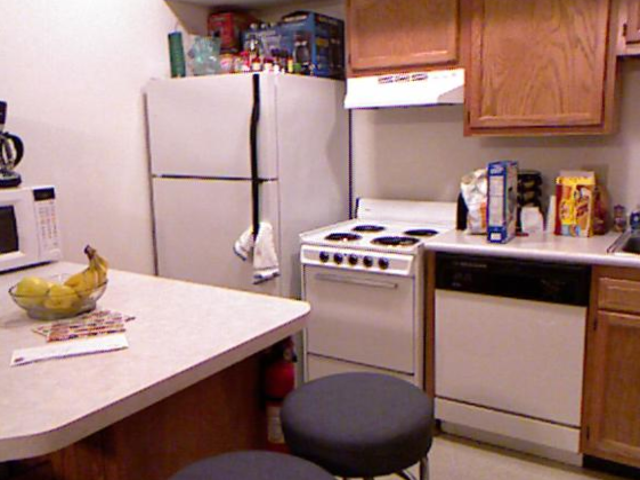} \ \
& \includegraphics[width=0.15\linewidth]{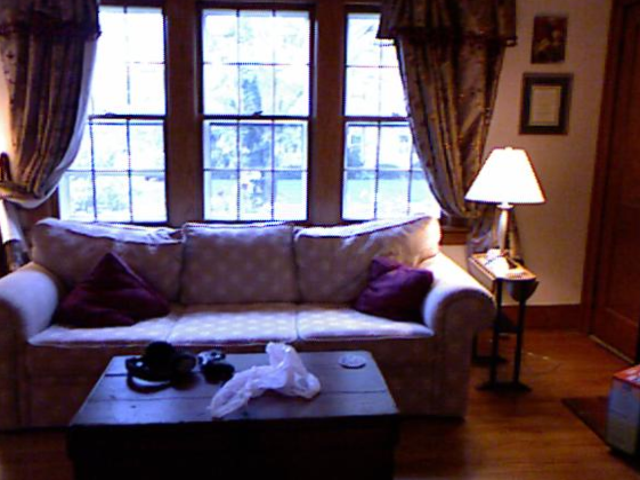} \ \
& \includegraphics[width=0.15\linewidth]{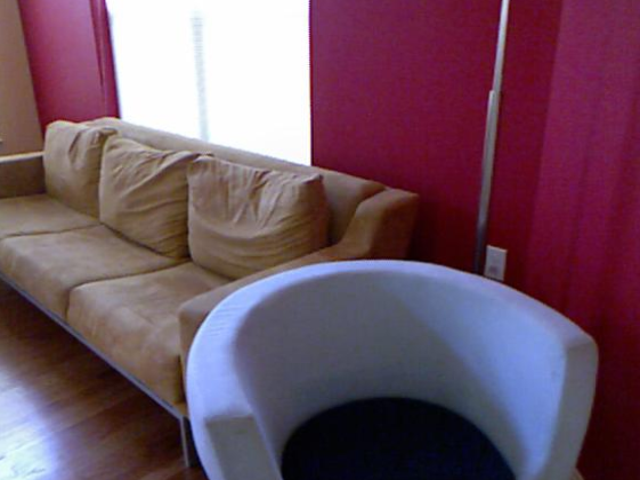} \\

(b) \ \
& \includegraphics[width=0.15\linewidth]{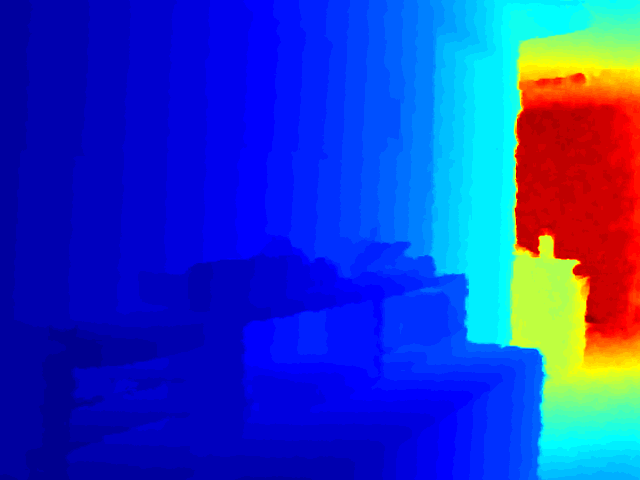} \ \
& \includegraphics[width=0.15\linewidth]{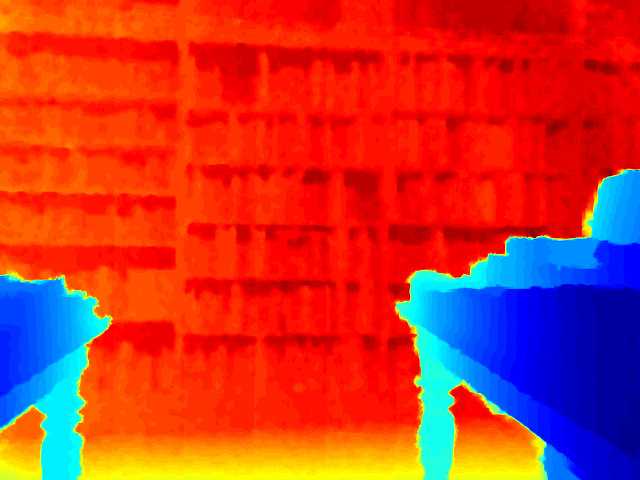} \ \
& \includegraphics[width=0.15\linewidth]{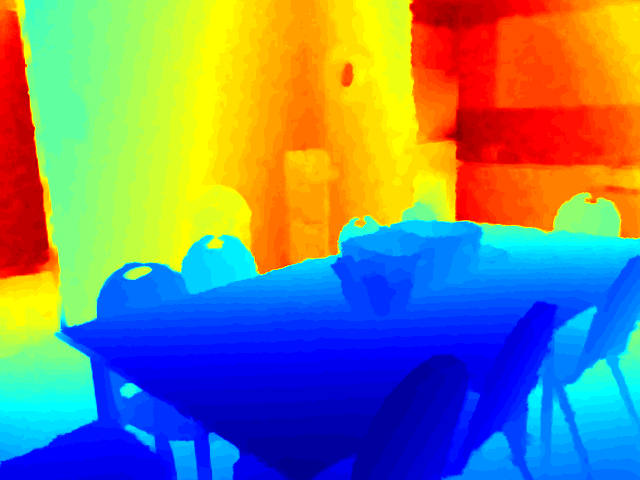} \ \
& \includegraphics[width=0.15\linewidth]{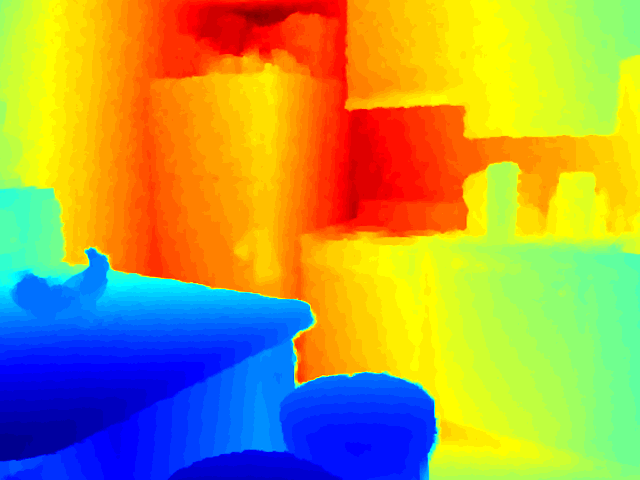} \ \
& \includegraphics[width=0.15\linewidth]{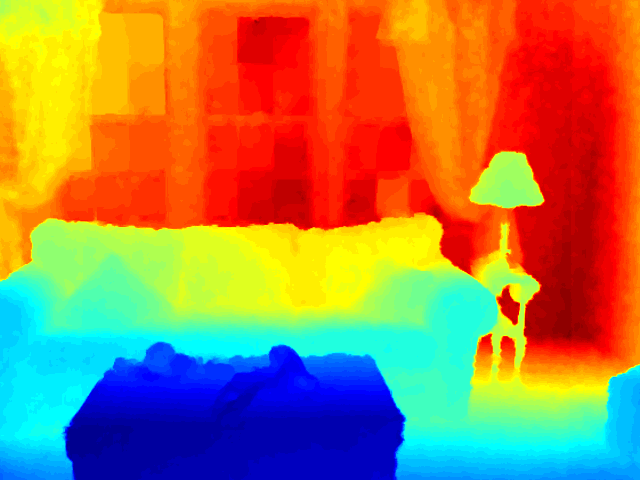} \ \
& \includegraphics[width=0.15\linewidth]{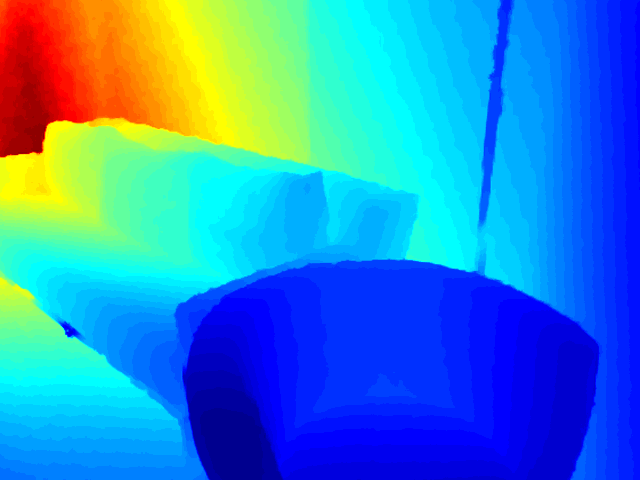} \\

(c) \ \
& \includegraphics[width=0.15\linewidth]{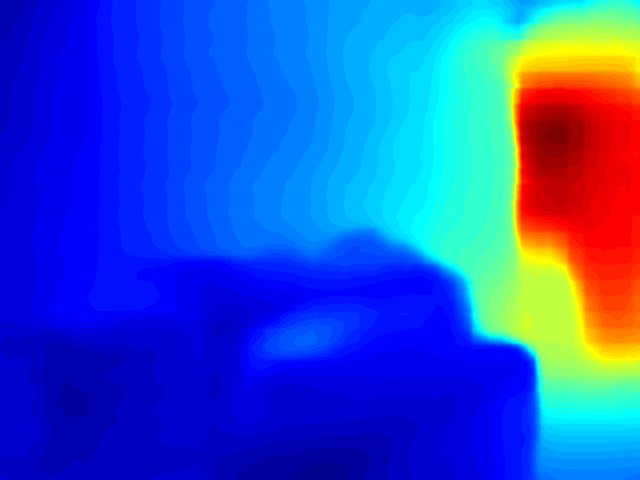} \ \
& \includegraphics[width=0.15\linewidth]{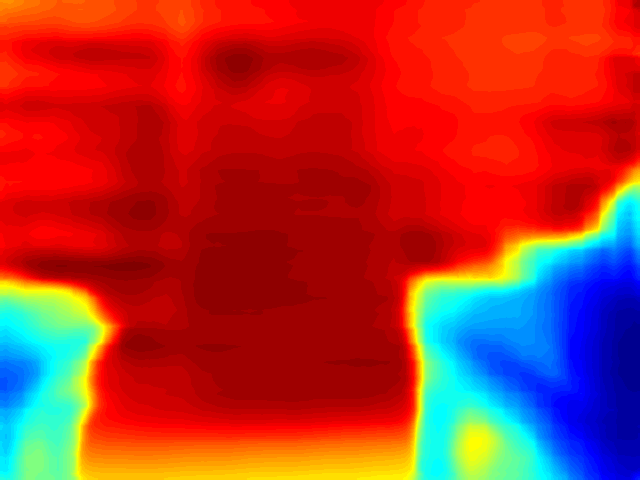} \ \
& \includegraphics[width=0.15\linewidth]{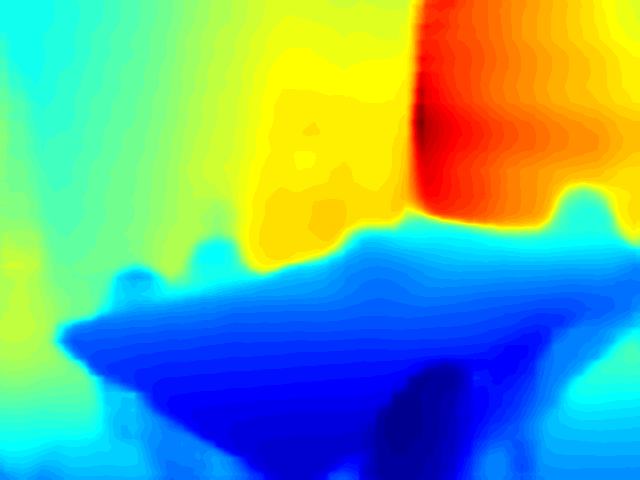} \ \
& \includegraphics[width=0.15\linewidth]{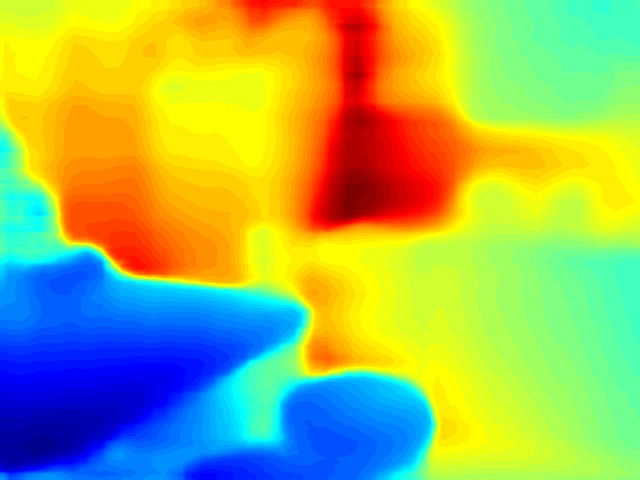} \ \
& \includegraphics[width=0.15\linewidth]{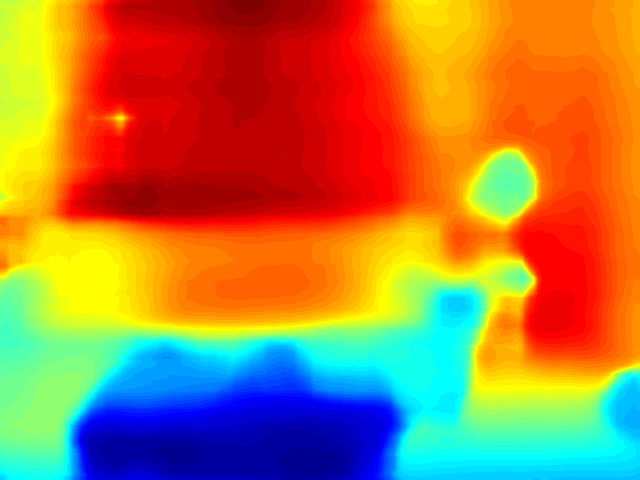} \ \
& \includegraphics[width=0.15\linewidth]{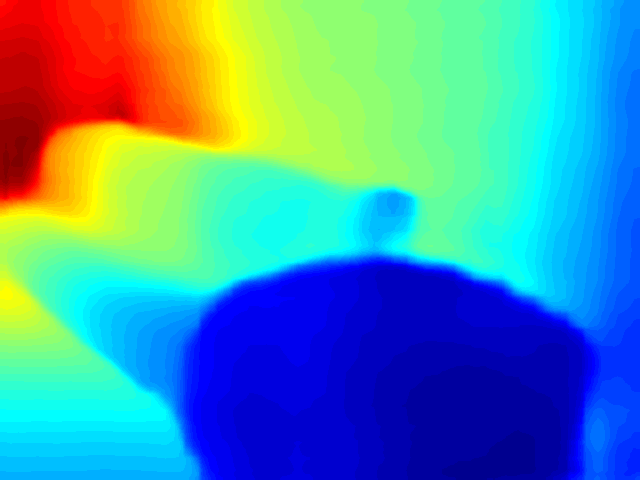} \\

(d) \ \
& \includegraphics[width=0.15\linewidth]{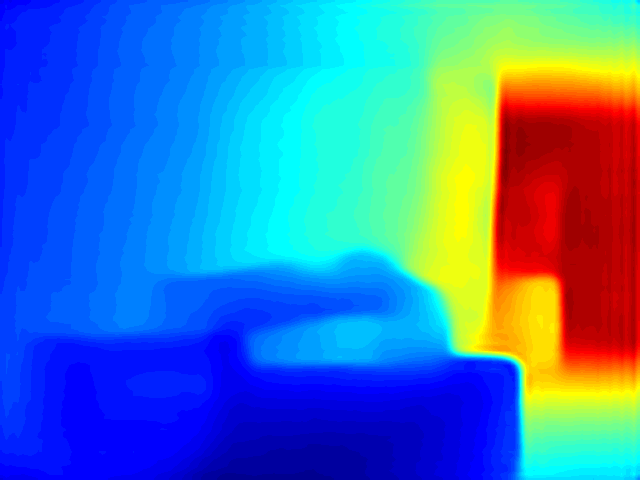} \ \
& \includegraphics[width=0.15\linewidth]{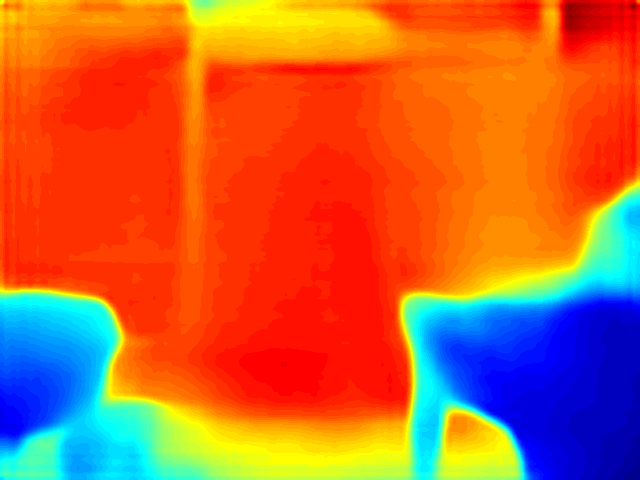} \ \
& \includegraphics[width=0.15\linewidth]{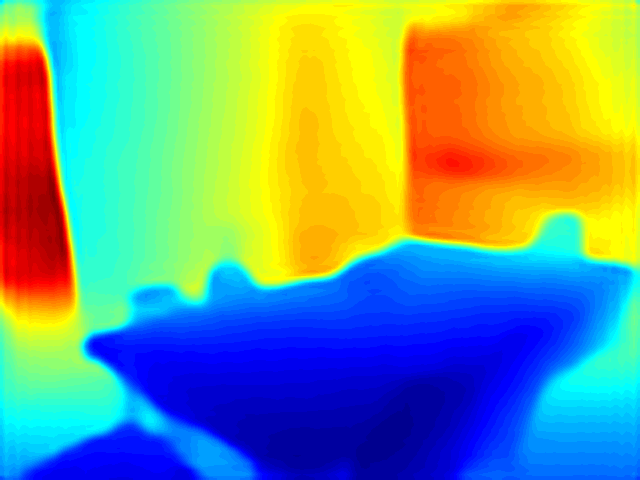} \ \
& \includegraphics[width=0.15\linewidth]{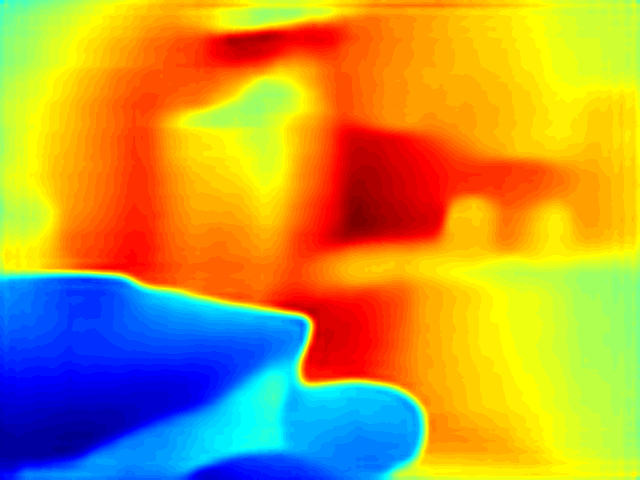} \ \
& \includegraphics[width=0.15\linewidth]{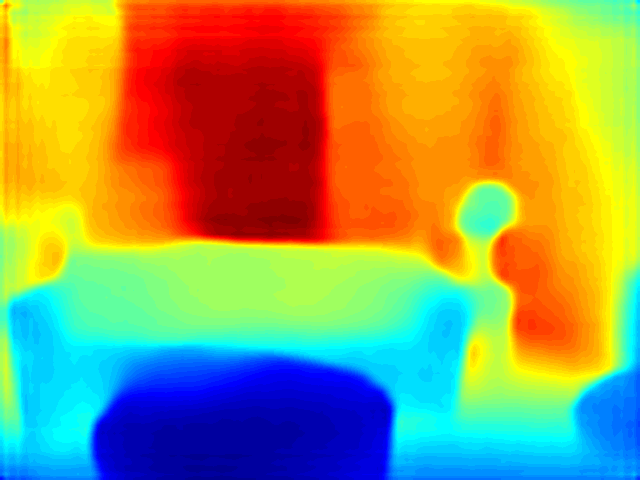} \ \
& \includegraphics[width=0.15\linewidth]{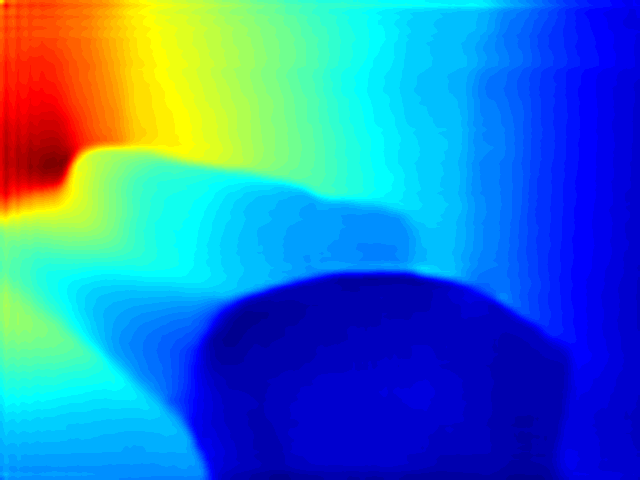} \\

(e) \ \
& \includegraphics[width=0.15\linewidth]{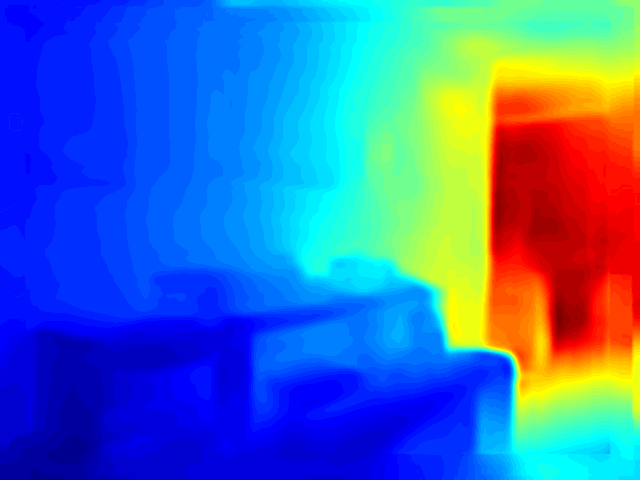} \ \
& \includegraphics[width=0.15\linewidth]{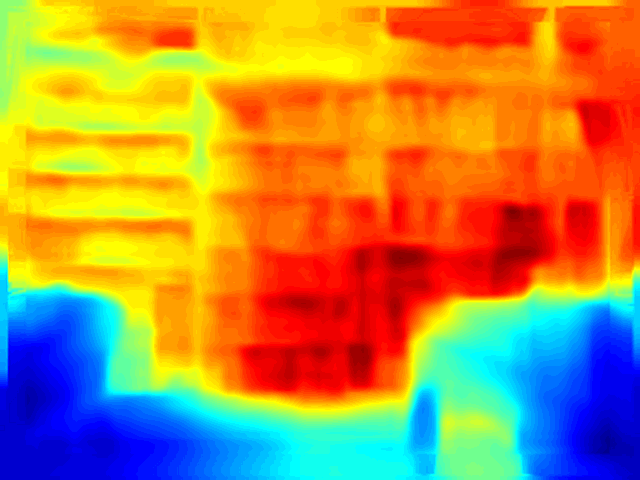} \ \
& \includegraphics[width=0.15\linewidth]{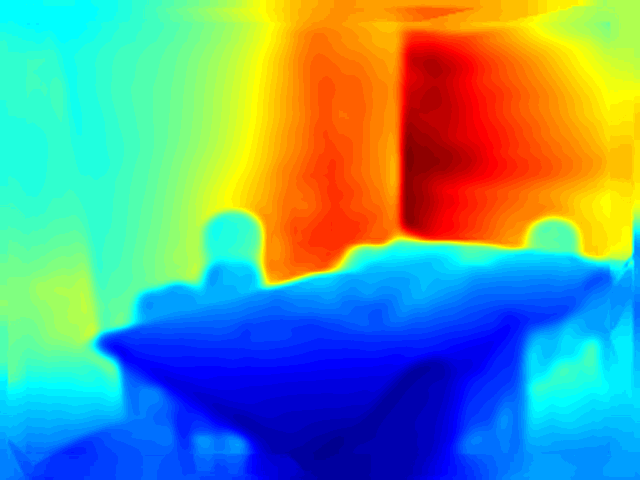} \ \
& \includegraphics[width=0.15\linewidth]{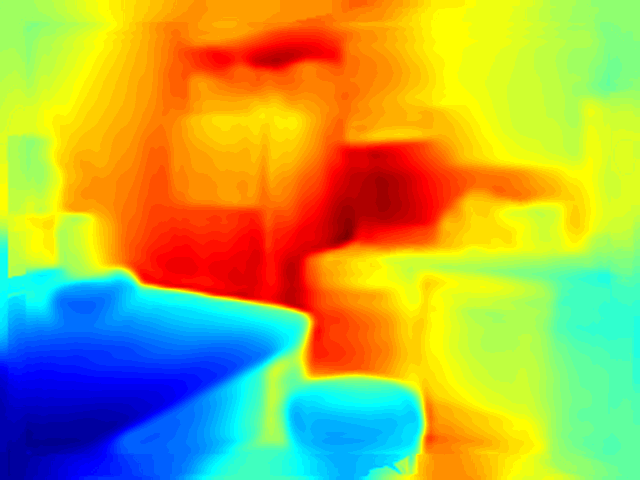} \ \
& \includegraphics[width=0.15\linewidth]{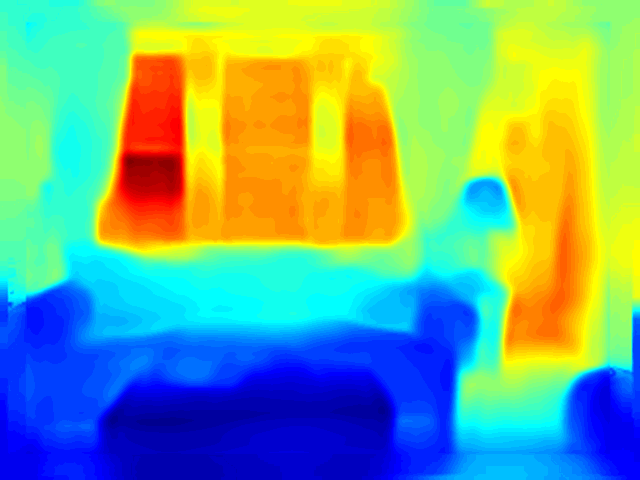} \ \
& \includegraphics[width=0.15\linewidth]{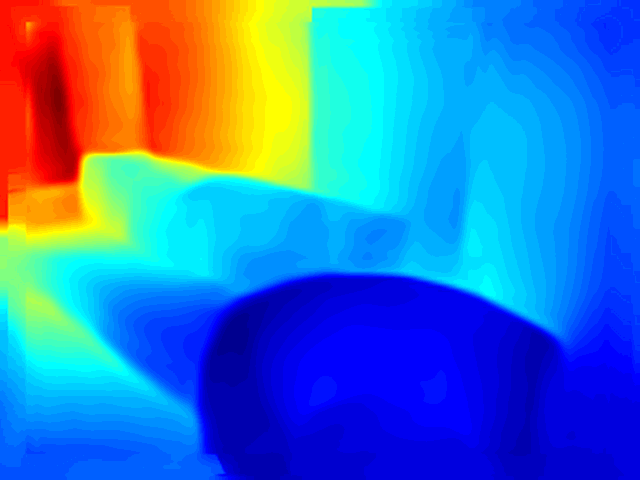} \\

\end{tabular}
}
\caption{Qualitative comparison of the estimated depth map on the NYU V2 dataset with our method and some state-of-the-art methods. Color indicates depth (red is far, blue is close). (a) RGB. (b) The Ground Truth. (c) Results of our proposed method. (d) Results of \cite{laina2016deeper}. (e) Results of \cite{eigen2015predicting} }
\label{fig:nyu2}
\end{figure*}

In Fig.~\ref{fig:nyu2}, we provide a qualitative comparison of our method with \cite{laina2016deeper} and \cite{eigen2014depth}. (We compare with these methods due to they published their results and they are the state-of-the-art methods at present). From Fig.\ref{fig:nyu2}, it is clear to observe that our results are of high visual quality, although we have not applied any post-processing methods.

\begin{table*}[htb]
\center
\caption{Depth estimation results on the NYU v2 dataset, $*$ represent the results are only calculated on the valuable pixels}
\resizebox{.9\linewidth}{!} {
\begin{tabular}{ | l | c | c  c  c | c  c  c |}
\hline
\multirow{3}{*}{{{Method}}} &\multirow{3}{*}{{{Train Num}}} &\multicolumn{3}{c|}{Accuracy} &\multicolumn{3}{c|}{Error}  \\
&{} &\multicolumn{3}{c|}{(higher is better)} &\multicolumn{3}{c|}{(lower is better)}\\
\cline{3-8}
&{}&$\delta < 1.25$ &$\delta < 1.25^2$ &$\delta < 1.25^3$ &Rel &log10 &Rms\\
\hline

Eigen \etal~\cite{eigen2015predicting}                     &{120K} &76.9\% &95.0\%  &98.8\% &0.158 &- &0.641\\
Cao \etal~\cite{cao2016estimating} &{120k} &80.0\% &95.6\% &98.8\% &0.148 &0.063 &{0.615}\\
Laina \etal~\cite{laina2016deeper}                      &{12k} &81.1\% &95.3\% &98.8\% &0.127 &0.055 &{0.573}\\
Xu \etal~\cite{Xu2017Multi} &{95k} &{81.1\%} &{95.4}\% &{98.7}\% &\bf{0.121} &\bf{0.052} &{0.586}\\
\hline
Ours                                            &{12k} &\bf{82.0\%} &\bf{96.0}\% &\bf{98.9}\% &{0.139} &{0.058} &\bf{0.505}\\
\hline
\end{tabular}
}
\label{tab:nyu2}
\end{table*}

\subsection{ KITTI dataset}
The KITTI dataset~\cite{geiger2012we} consists of a large number of outdoor street scene images of the resolution $376 \times 1242$. We utilized the ``Eigen'' training/testing split, which consists of 22600 training images and 697 testing images. We fill in the invalid pixel of the raw depth map with the ``colorization'' method, which is provided in the toolbox of NYU V2 dataset~\cite{Silberman2012Indoor}. For the error calculation, We only consider the lower crop of the image of size $256 \times 1242$. While in the training phase, we input the entire image to the network for more context information. We compare with the state-of-the-art methods Eigen \etal~\cite{eigen2014depth}, Garg \etal~\cite{Garg2016Unsupervised} and Godard \etal~\cite{Godard2017Unsupervised}.

The original image resolution is $376 \times 1240$. We downsampled the images to $188 \times 620$ as our network input. The resolution of the our network output is $94 \times 310$, which is half of the input size. For this dataset, we quantize the depth value into 50 bins. 

In Table~\ref{tab:KITTI}, we compared our method with the recent published state-of-the-art methods~\cite{eigen2014depth,Garg2016Unsupervised,Godard2017Unsupervised}.

\begin{figure*}[htb]
\centering
\scalebox{0.9}
{
\begin{tabular}{@{}c@{}c@{}c@{}c} 
(a)\ \
&\includegraphics[width=0.3\linewidth]{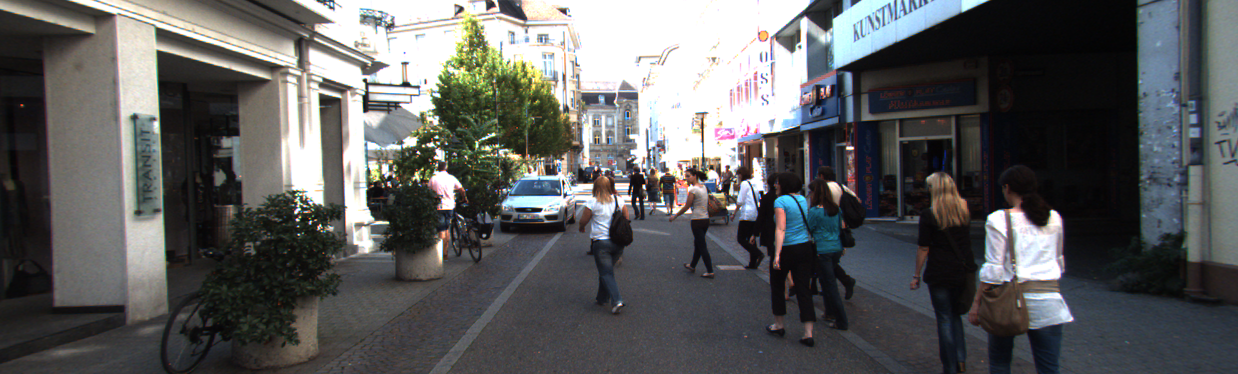} \ \
&\includegraphics[width=0.3\linewidth]{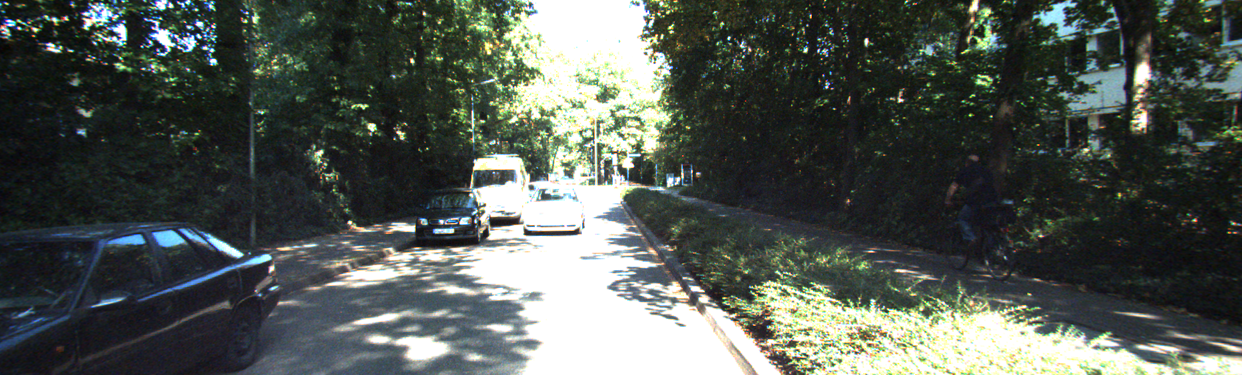} \ \
&\includegraphics[width=0.3\linewidth]{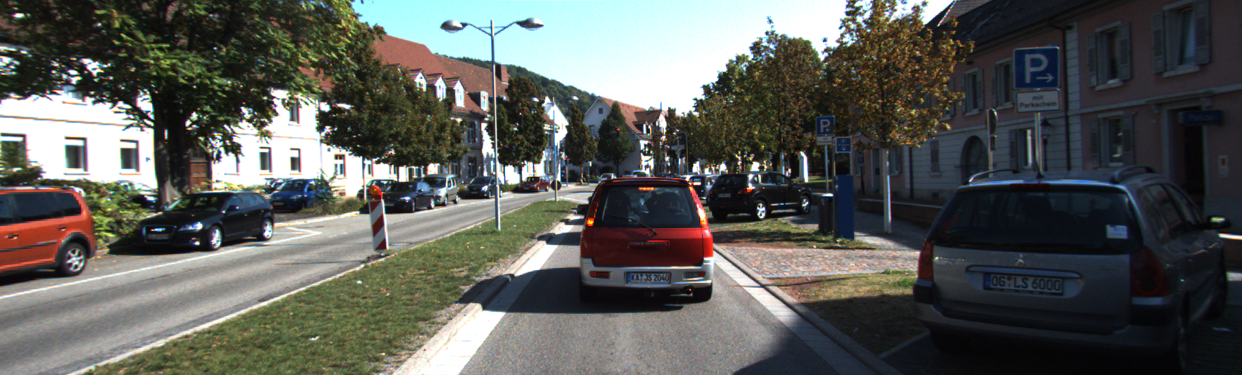} \\
(b)\ \
&\includegraphics[width=0.3\linewidth]{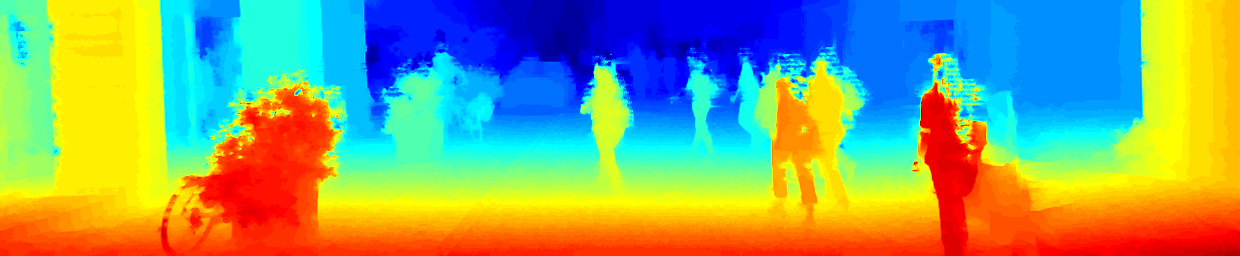} \ \
&\includegraphics[width=0.3\linewidth]{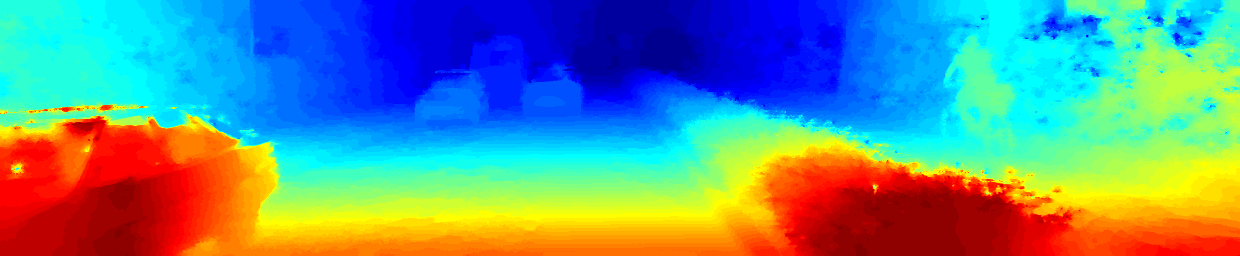} \ \
&\includegraphics[width=0.3\linewidth]{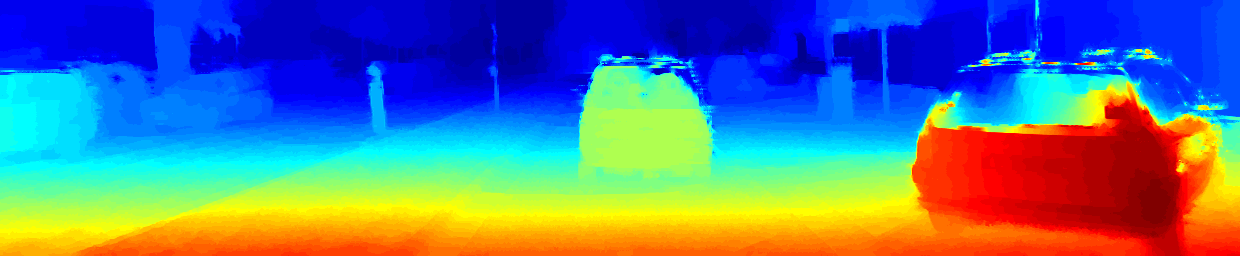} \\
(c)\ \
&\includegraphics[width=0.3\linewidth]{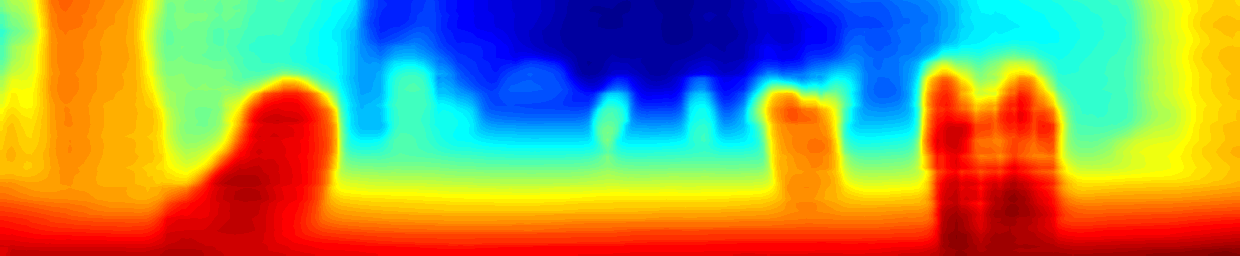} \ \
&\includegraphics[width=0.3\linewidth]{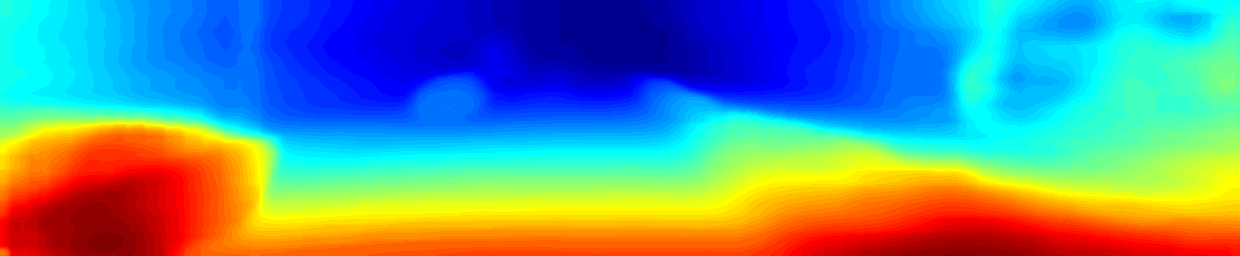} \ \
&\includegraphics[width=0.3\linewidth]{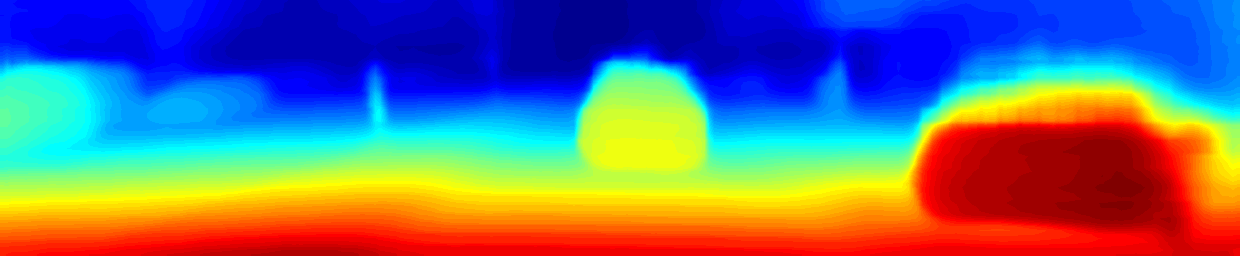} \\
(d)\ \
&\includegraphics[width=0.3\linewidth]{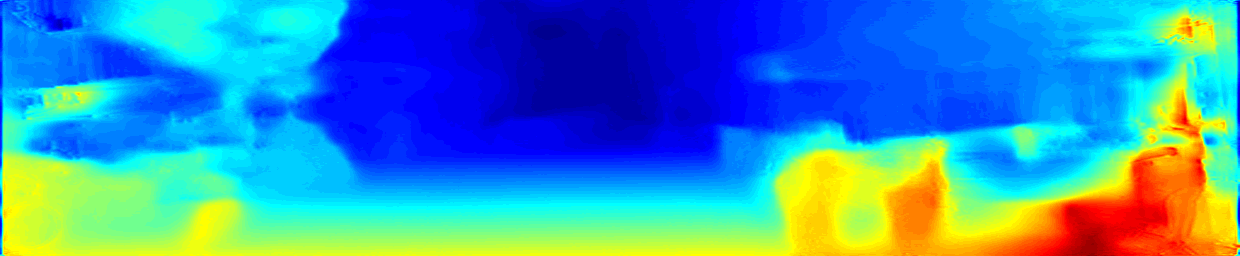} \ \
&\includegraphics[width=0.3\linewidth]{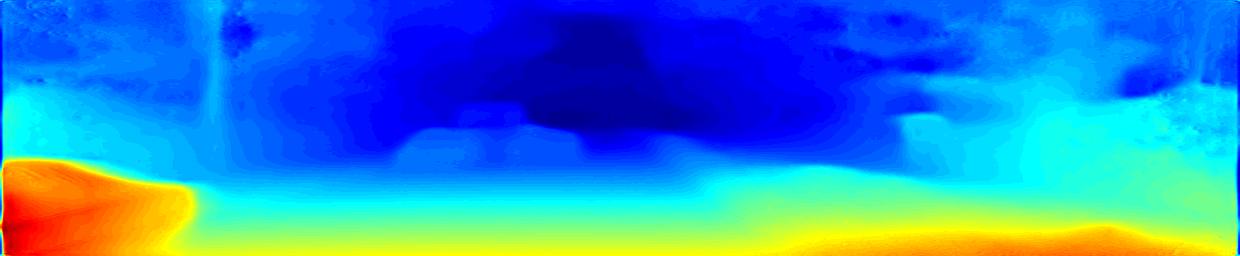} \ \
&\includegraphics[width=0.3\linewidth]{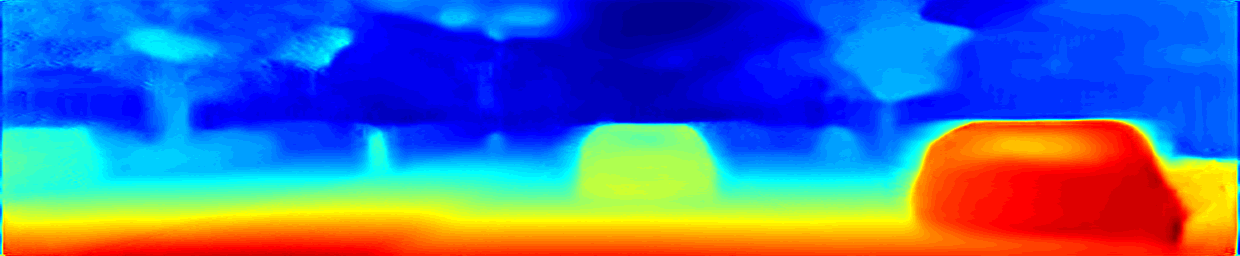} \\
(e)\ \
&\includegraphics[width=0.3\linewidth]{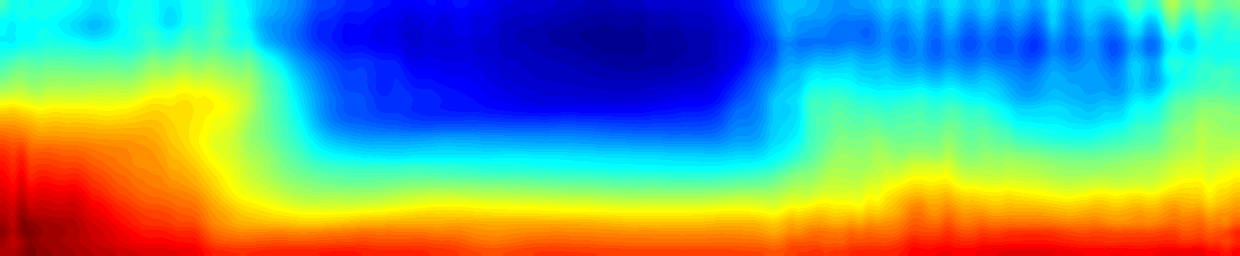} \ \
&\includegraphics[width=0.3\linewidth]{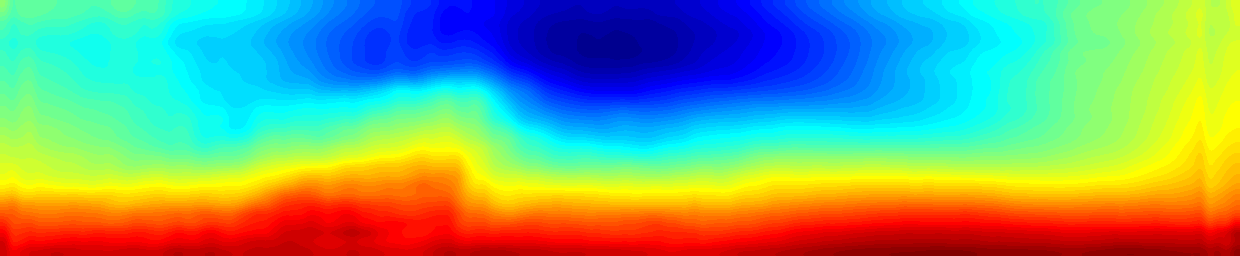} \ \
&\includegraphics[width=0.3\linewidth]{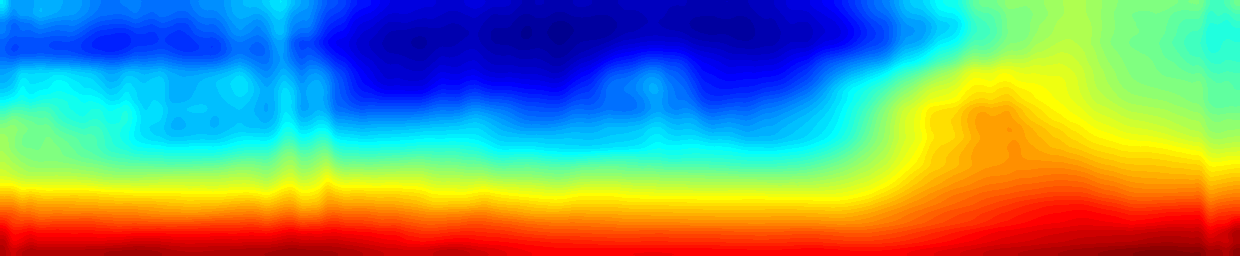} \\
\end{tabular}
}
\caption{Qualitative comparison of the depth map estimated on KITTI dataset. Color indicates depth (red is close, blue is far). (a) RGB. (b) The Ground Truth. (c) Results of proposed method. (d) Results of \cite{Godard2017Unsupervised}. (e) Results of \cite{eigen2015predicting}}
\label{fig:kitti}
\end{figure*}

In Fig.~\ref{fig:kitti}, we provide a qualitative comparison of our method with \cite{eigen2015predicting} and \cite{Godard2017Unsupervised}. (We compare with these methods due to they published their results and they are the state-of-the-art methods at present). From Fig.\ref{fig:kitti}, it is clear to observe that our results are of high visual quality, although we have not applied any post-processing methods.

\begin{table*}[htb]
\caption{Depth estimation results on the KITTI dataset.}
\center
\resizebox{.9\linewidth}{!} {
\begin{tabular}{ | l | c | c  c  c | c  c  c |}
\hline
\multirow{3}{*}{{{Method}}} &\multirow{3}{*}{{{Train Num}}} &\multicolumn{3}{c|}{Accuracy} &\multicolumn{3}{c|}{Error}  \\
&{} &\multicolumn{3}{c|}{(higher is better)} &\multicolumn{3}{c|}{(lower is better)}  \\
\cline{3-8}
&{}&$\delta < 1.25$ &$\delta < 1.25^2$ &$\delta < 1.25^3$ &Rel &log10 &Rms  \\
\hline
Eigen \etal~\cite{eigen2014depth} &22600 &70.2\% &89.0\% &95.8\% &0.203 &- &6.307\\
Godard \etal~\cite{Godard2017Unsupervised} &22600 &80.3\% &92.2\% &96.4\% &0.148 &- &5.927\\
Godard \etal~\cite{Godard2017Unsupervised} cap 50m &22600 &81.8\% &93.1\% &96.9\% &0.140 &- &4.471\\
\hline
Ours                                        &22600  &\bf{85.6\%} &\bf{96.2\%} &\bf{98.8\%} &\bf{0.113} &\bf{0.049} &\bf{4.687}\\
Ours cap 50m &22600  &\bf{86.4\%} &\bf{96.4\%} &\bf{98.9\%} &\bf{0.109} &\bf{0.047} &\bf{3.906}\\
\hline
\end{tabular}
}
\label{tab:KITTI}
\end{table*}

%
\section{Performance Analysis}\label{more analysis}
In this section, we present more analysis of our model, where the experiments are conducted on the NYU V2 dataset. First, we present a component analysis of our network architecture design, \ie, the contribution of each component. Second, we analyze the distribution of our network output, which demonstrates the necessary of our soft-weighted-sum inference strategy. 
\subsection{Effect of Architecture Design}
In order to explore the effectiveness of our hierarchical fusion dilated CNN, we conduct the following component analyze experiments. First, we utilize the normal convolution kernel instead of the dilated convolution kernel in the last 2 parts of our network. Second, we remove the skip connection structure. At last, we use the network without dilated convolution and skip connection. The corresponding experimental results are presented in Tab.~\ref{tab:architecture}. As we can see, both  dilated convolution and hierarchical fusion play important roles in achieving improved performance.

\begin{table}[!htb]
\center
\caption{Component evaluation for our CNN architecture design and soft-weighted sum inference.}
\resizebox{0.9\linewidth}{!}
{
\begin{tabular}{ | l | c  c  c | c  c  c |}
\hline
\multirow{2}{*}{{{method}}}                &\multicolumn{3}{c|}{Accuracy (\%)} &\multicolumn{3}{c|}{Error}  \\
\cline{2-7}
                                           &$\delta < 1.25$ &$\delta < 1.25^2$ &$\delta < 1.25^3$ &Rel &log10 &Rms\\
\hline
no dilation                &{78.02\%} &{94.61\%} &{98.52\%} &{0.157} &{0.066} &{0.559}\\
no concat layer        &{81.64\%} &{95.9\%} &{98.8\%} &{0.141} &{0.059} &{0.509}\\
ours hard-max                               &{81.82\%} &{95.53\%} &{98.53\%} &{0.142} &{0.060} &{0.531}\\
\hline
ours soft-weighted sum                               &\bf{82.0\%} &\bf{96.0}\% &\bf{98.9}\% &\bf{0.139} &\bf{0.058} &\bf{0.505}\\
\hline
\end{tabular}}
\label{tab:architecture}
\end{table}

\subsection{Effect of Soft-Weighted-Sum Inference}
One important contribution of this work is the proposed soft-weighted-sum inference. Here, we would like to elaborate the necessity and effectiveness of it. 

\begin{figure*}[htb]
\centering
\scalebox{.9}
{
\begin{tabular}{@{}c@{}c@{}c@{}c@{}c@{}c@{}c}

& \includegraphics[width=0.1825\linewidth]{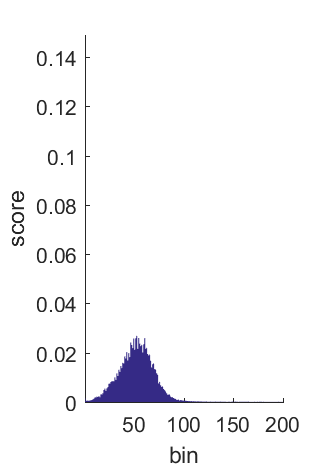} \ \
& \includegraphics[width=0.15\linewidth]{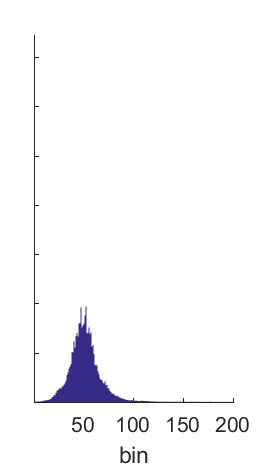} \ \
& \includegraphics[width=0.15\linewidth]{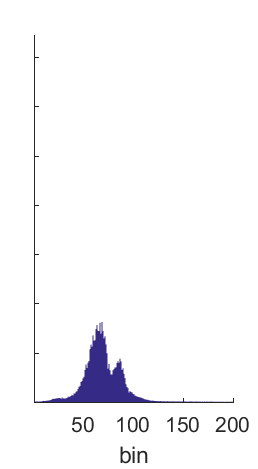} \ \
& \includegraphics[width=0.15\linewidth]{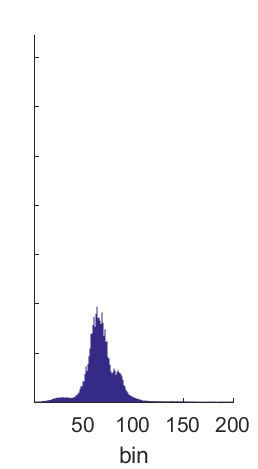} \ \
& \includegraphics[width=0.15\linewidth]{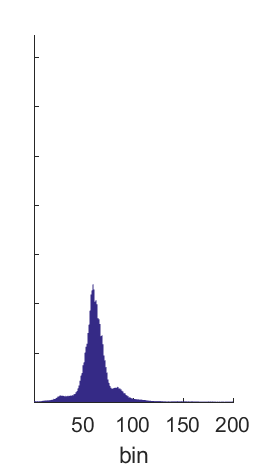} \ \
& \includegraphics[width=0.15\linewidth]{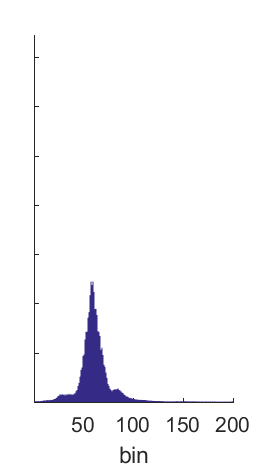} \\

& \includegraphics[width=0.1825\linewidth]{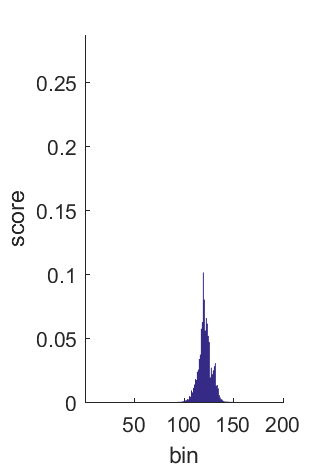} \ \
& \includegraphics[width=0.15\linewidth]{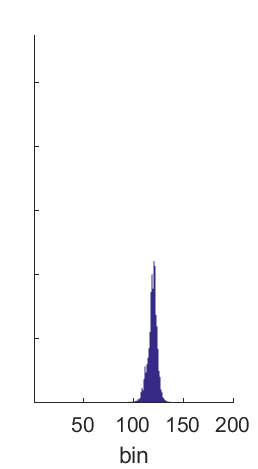} \ \
& \includegraphics[width=0.15\linewidth]{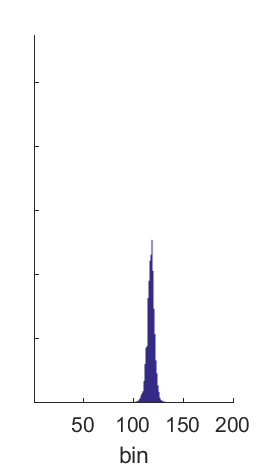} \ \
& \includegraphics[width=0.15\linewidth]{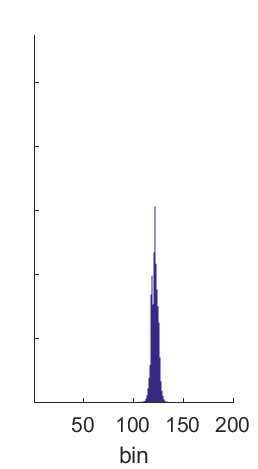} \ \
& \includegraphics[width=0.15\linewidth]{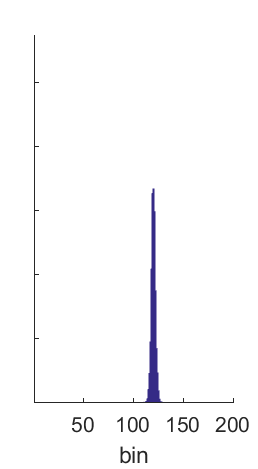} \ \
& \includegraphics[width=0.15\linewidth]{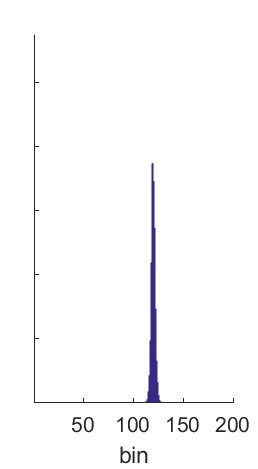} \\

& \includegraphics[width=0.1825\linewidth]{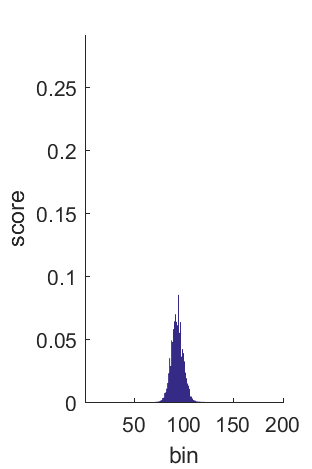} \ \
& \includegraphics[width=0.15\linewidth]{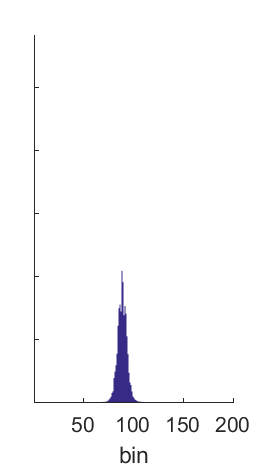} \ \
& \includegraphics[width=0.15\linewidth]{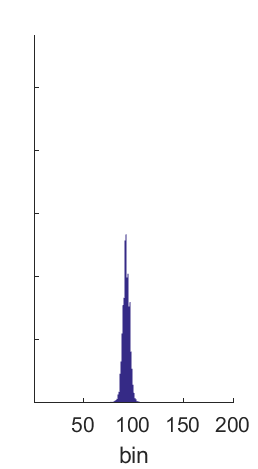} \ \
& \includegraphics[width=0.15\linewidth]{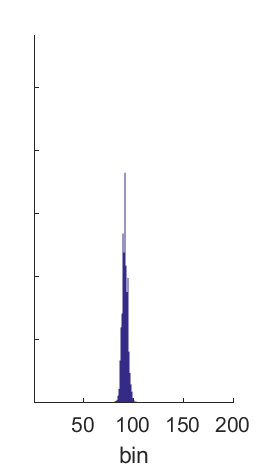} \ \
& \includegraphics[width=0.15\linewidth]{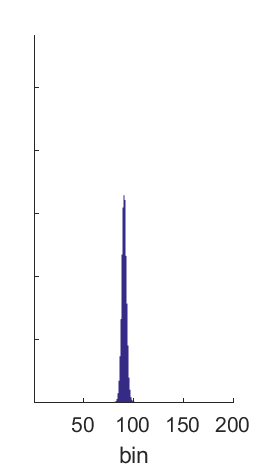} \ \
& \includegraphics[width=0.15\linewidth]{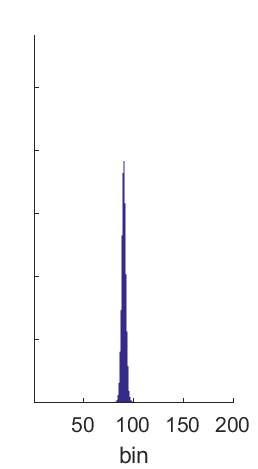} \\

&iteration 5000\ \
&10000\ \
& 20000\ \
& 30000\ \
& 40000\ \
& 50000\\
\end{tabular}
}
\caption{Typical score distribution variation of our network output. The points are randomly selected from NYU2 dataset.}
\label{fig:curve}
\end{figure*}

Firstly, we give the probability distribution variation of randomly selected positions along the training in Fig.~\ref{fig:curve}. The most interesting thing is that: In the training phase, we utilize the multinomial logistic loss, which means we don't specially discriminate the distance between the ``nearby'' and ``further'' classes. While, the probability distribution is rather clustered. More interestingly, the probability distribution roughly follow the Gaussian distribution, which means it is symmetric. At last, as the training goes on, the distribution of probability is becoming more concentrated, but always maintains symmetry similar to that of Gaussian distribution. 

\begin{figure}[ht]
\centering{\includegraphics[width=0.9\linewidth]{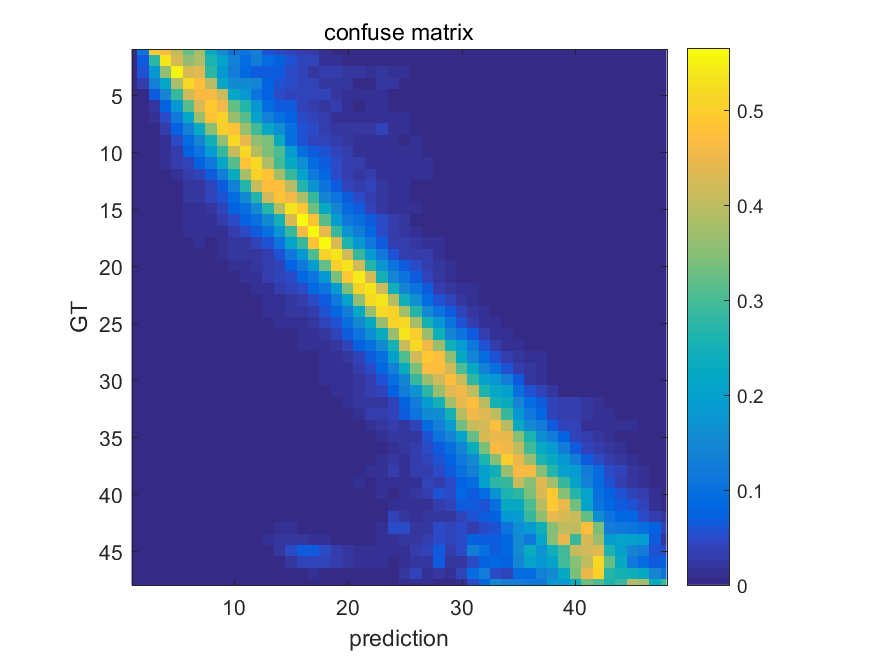}}
\caption{Confusion matrix on the NYU2 dataset. Here, we merge the 200 bins to 50 for better illustration.}
\label{fig:confuse_matrix}
\end{figure}

Secondly, we use the hard-max inference and give the confusion matrix in Fig. \ref{fig:confuse_matrix}. The confusion matrix presents a kind of diagonal dominant and symmetric structure, which means most of the error prediction occurs in nearby classes.

Thirdly, we increase the number of depth bins. Under the same training setting, we present the variation of ``pixel accuracy'' and the relative errors in Tab.~\ref{tab:num bins}. With the increase of number of bins, the ``pixel accuracy'' drop dramatically, while the relative error keeps stable. This trend presents that: 1) At present, the network cannot distinguish the very detailed distance variation even we train it with the very detailed ``label''. In other words, ``depth perception'' ability of the network is limited.


\begin{table}[htb]
\center
\caption{Pixel-wise accuracy and Rel w.r.t. number of bins.}
\resizebox{.9\linewidth}{!}
{
\begin{tabular}{ | l | c | c | c | c | c |}
\hline
num of bins              &50 &100 &200 &500 &1000\\
\hline
pixel accuracy (\%)           &67   &41  &25 &12  &7 \\
Rel                      &0.182   &0.142  &0.139  &0.138  &0.140 \\
\hline
\end{tabular}
}
\label{tab:num bins}
\end{table}

%
\section{Conclusions}
In this paper, we have proposed a deep end-to-end classification based framework to monocular depth estimation. By using both dilated convolution and hierarchical fusion of multi-scale features, our framework is able to deal with the real world difficulties in multi-scale depth estimation. Extensive experiments on both indoor and outdoor benchmarking datasets show the superiority of our method compared with the current state-of-the-art methods. More importantly, experiments also demonstrate that our model is able to learn a probability distribution among different depth labels, which inspires the proposed soft-weighted-sum inference.

\section*{Acknowledgments}
This work was supported in part by Australian Research Council (ARC) grants (DE140100180, DP120103896, LP100100588, CE140100016), Australia ARC Centre of Excellence Program on Roboitic Vision, NICTA (Data61) and Natural Science Foundation of China (61420106007). 

\bibliography{CSRef}

\end{document}